\documentclass{article}

    \PassOptionsToPackage{numbers, compress}{natbib}
\usepackage[final]{neurips_2021}

\usepackage[utf8]{inputenc} %
\usepackage[T1]{fontenc}    %

\usepackage{amsmath,amsfonts,bm}

\def\eqref#1{equation~\ref{#1}}

\def\1{\bm{1}}

\def\vs{{\bm{s}}}

\DeclareMathAlphabet{\mathsfit}{\encodingdefault}{\sfdefault}{m}{sl}
\SetMathAlphabet{\mathsfit}{bold}{\encodingdefault}{\sfdefault}{bx}{n}

\usepackage{color,xcolor}
\usepackage{epsfig}
\usepackage{graphicx}

\usepackage{adjustbox}
\usepackage{array}
\usepackage{booktabs}
\usepackage{colortbl}
\usepackage{wrapfig}
\usepackage{hhline}
\usepackage{multirow}
\usepackage{subcaption} %
\usepackage{wrapfig}
\usepackage{floatflt}

\usepackage{amsmath,amsfonts,amssymb}
\usepackage{mathtools}  %
\usepackage{bm}
\usepackage{nicefrac}
\usepackage{microtype}

\usepackage{changepage}
\usepackage{extramarks}
\usepackage{fancyhdr}
\usepackage{lastpage}
\usepackage{setspace}
\usepackage{soul}
\usepackage{xspace}

\usepackage[pagebackref=true,breaklinks=true,colorlinks,citecolor=gray]{hyperref}

\usepackage{url}

\usepackage{enumerate}
\usepackage{todonotes} %
\usepackage{enumitem}  %

\usepackage{titlesec}

\usepackage{makecell}

\usepackage{pifont} %
\usepackage{subcaption}

\usepackage{algorithm,algpseudocode}

\usepackage{amsthm}
\newcommand{\stderr}[1]{\scriptsize $\pm #1$}

\newcolumntype{L}[1]{>{\raggedright\let\newline\\\arraybackslash\hspace{0pt}}m{#1}}
\newcolumntype{C}[1]{>{\centering\let\newline\\\arraybackslash\hspace{0pt}}m{#1}}
\newcolumntype{R}[1]{>{\raggedleft\let\newline\\\arraybackslash\hspace{0pt}}m{#1}}

\newcommand{\sectapp}[1]{Appendix~\ref{#1}}

\newcommand{\fig}[1]{Fig.~\ref{#1}}
\newcommand{\tbl}[1]{Table~\ref{#1}}
\newcommand{\alg}[1]{Algorithm.~\ref{#1}}

\newcommand{\ignore}[1]{}

\makeatletter
\DeclareRobustCommand\onedot{\futurelet\@let@token\@onedot}
\def\@onedot{\ifx\@let@token.\else.\null\fi\xspace}

\def\eg{e.g\onedot} 
\def\ie{i.e\onedot}

\def\vs{\emph{vs}\onedot}
\def\wrt{w.r.t\onedot}

\def\etal{et al\onedot}

\makeatother

\definecolor{MyDarkBlue}{rgb}{0,0.08,1}
\definecolor{MyDarkGreen}{rgb}{0.02,0.6,0.02}
\definecolor{MyDarkRed}{rgb}{0.8,0.02,0.02}
\definecolor{MyDarkOrange}{rgb}{0.40,0.2,0.02}
\definecolor{MyPurple}{RGB}{111,0,255}
\definecolor{MyRed}{rgb}{1.0,0.0,0.0}
\definecolor{MyGold}{rgb}{0.75,0.6,0.12}
\definecolor{MyDarkgray}{rgb}{0.66, 0.66, 0.66}

\newcommand{\model}{G2L2\xspace}
\newcommand{\modelfull}{Grammar-Based Grounded Lexicon Learning\xspace}
\newcommand{\ckyee}{CKY-E2\xspace}

\newcommand{\myparagraph}[1]{\noindent\textbf{#1}}

\newcommand{\mycell}[1]{\begin{tabular}[t]{@{}l@{}l}#1\end{tabular}}

\title{Grammar-Based Grounded Lexicon Learning}

\author{%
  Jiayuan Mao\\
  MIT
  \And
  Haoyue Shi\\
  TTIC
  \And
  Jiajun Wu\\
  Stanford University
  \And
  Roger P. Levy\\
  MIT
  \And
  Joshua B. Tenenbaum\\
  MIT
}

\begin{document}

{
\setlength{\tabcolsep}{5pt}
\maketitle
}

\footnotetext{Correspondence to Jiayuan Mao: jiayuanm@mit.edu. Project page: \url{http://g2l2.csail.mit.edu}.}

\begin{abstract}
We present \modelfull (\model), a lexicalist approach toward learning a compositional and grounded meaning representation of language from grounded data, such as paired images and texts. At the core of \model is a collection of lexicon entries, which map each word to a tuple of a syntactic type and a neuro-symbolic semantic program. For example, the word {\it shiny} has a syntactic type of \emph{adjective}; its neuro-symbolic semantic program has the {\it symbolic} form $\lambda x.\textit{filter}(x, \textbf{SHINY})$, where the concept {\bf SHINY} is associated with a {\it neural network} embedding, which will be used to classify shiny objects. Given an input sentence, \model first looks up the lexicon entries associated with each token. It then derives the meaning of the sentence as an executable neuro-symbolic program by composing lexical meanings based on syntax. The recovered meaning programs can be executed on grounded inputs. To facilitate learning in an exponentially-growing compositional space, we introduce a joint parsing and expected execution algorithm, which does local marginalization over derivations to reduce the training time. We evaluate \model on two domains: visual reasoning and language-driven navigation. Results show that \model can generalize from small amounts of data to novel compositions of words.
\end{abstract}
\vspace{-10pt}
\section{Introduction}
\vspace{0pt}

Human language learning suggests several desiderata for machines learning from language. Humans can learn grounded and compositional representations for novel words from few examples. These representations are grounded on contexts, such as visual perception. We also know how these words relate with each other in composing the meaning of a sentence. \emph{Syntax}---the structured, order-sensitive relations among words in a sentence---is crucial in humans' learning and compositional abilities for language.
According to \emph{lexicalist} linguistic theories \cite{pollard-sag:1994,steedman:2000,bresnan-etal:2016-lexical}, syntactic knowledge involves a small number of highly abstract and potentially universal combinatory rules, together with a large amount of learned information in the lexicon: a rich syntactic type and meaning representation for each word.

\begin{figure*}
    \centering
    \includegraphics[width=\textwidth]{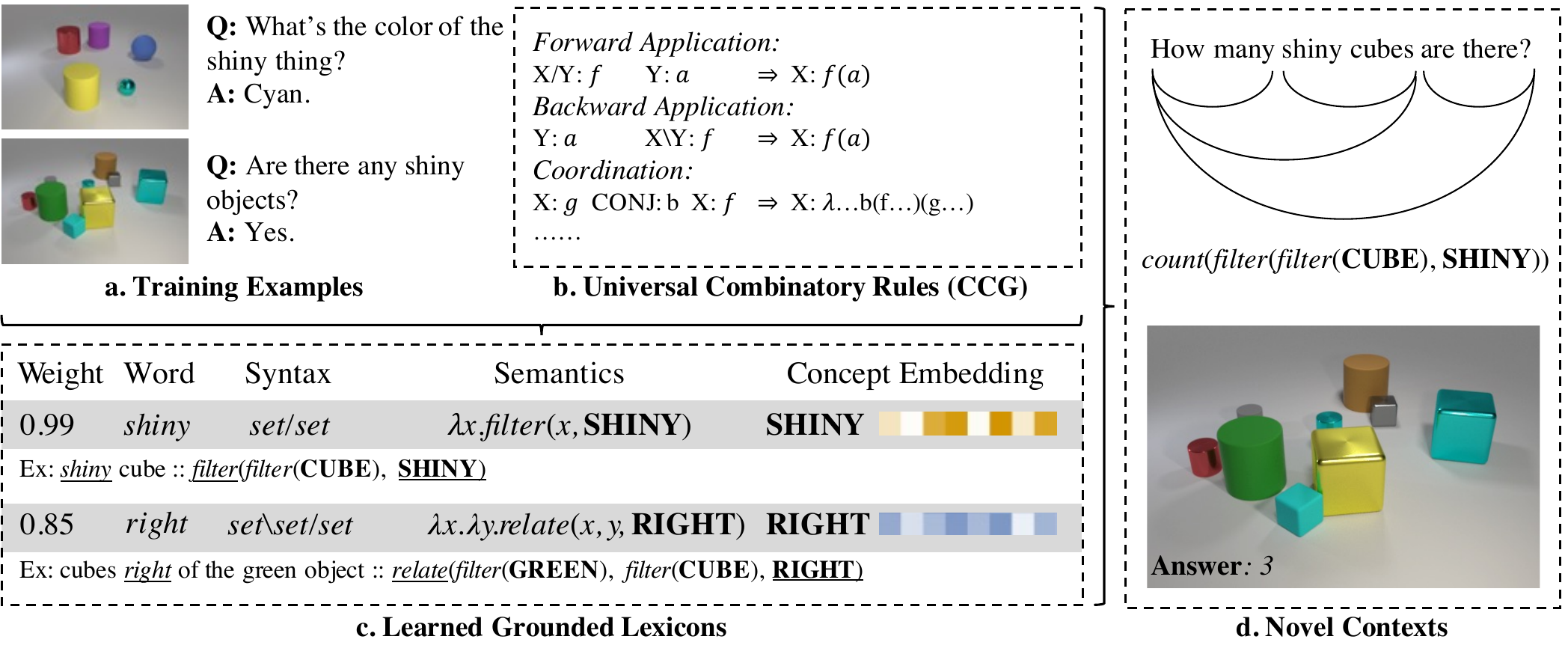}
    \vspace{-2em}
    \caption{\model learns from grounded language data, for example, by looking at images and reading parallel question--answer pairs. It learns a collection of grounded lexicon entries comprised of weights, syntax types, semantics forms, and optionally, grounded embeddings associated with semantic concepts. These lexicon entries can be used to parse questions into programs.}
    \vspace{-1.5em}
    \label{fig:teaser}
\end{figure*}

\fig{fig:teaser} illustrates this idea in a visually grounded language acquisition setup. The language learner looks at a few examples containing the novel word {\it shiny} (\fig{fig:teaser}a). They also have a built-in, compact but universal set of combinatory grammar rules (\fig{fig:teaser}b) that describes how the semantic program of words can be combined based on their syntactic types. The learner can recover the syntactic type of the novel word and its semantic meaining. For example, {\it shiny} is an adjective and its meaning can be grounded on visually shiny objects in images (\fig{fig:teaser}c). This representation supports the interpretation of novel sentences in a novel visual context (\fig{fig:teaser}d).

In this paper, we present \modelfull (\model), a neuro-symbolic framework for grounded language acquisition. At the core of \model is a collection of grounded lexicon entries. Each lexicon entry maps a word to (i) a syntactic type, and (ii) a neuro-symbolic semantic program. For example, the lexicon entry for the English word {\it shiny} has a syntactic type of {\it objset/objset}: it will compose with another constituent of type {\it objset} on its right, and produces a new constituent of syntactic type {\it objset}. For example, in \fig{fig:teaser}d, the word {\it shiny} composes with the word {\it cube} and yields a new constituent of type {\it objset}. The neuro-symbolic semantic program for {\it shiny} has the form $\lambda x.\textit{filter}(x, \textbf{SHINY})$, where $\textbf{SHINY}$ is a concept  {\it automatically} discovered by \model and associated with a learned vector embedding for classifying shiny objects. \model parses sentences based on these grounded lexicon entries and a small set of combinatory categorial grammar~\cite[CCG; ][]{steedman1996surface} rules. Given an input question, \model will lookup the lexicon entries associated with each token, and compose these lexical semantic programs based on their syntactic types. 

\model takes a lexicalist approach toward grounded language learning and focuses on data efficiency and compositional generalization to novel contexts. Inspired by lexicalist linguistic theories, but in contrast to neural network-based end-to-end learning, \model uses a compact symbolic grammar to constraint how semantic programs of individual words can be composed, and focuses on learning the lexical representation. This approach brings us strong data efficiency in learning new words, and strong generalization to new word compositions and sentences with more complex structures.

We are interested in jointly learning these neuro-symbolic grounded lexicon entries and the grounding of individual concepts from grounded language data, such as by simultaneously looking at images and reading parallel question--answer pairs. This is particularly challenging because the number of candidate lexicon entry combinations of a sentence grows exponentially with respect to the sentence length. For this reason, previous approaches to lexicon learning have either assumed an expert-annotated set of lexical entries~\cite{Zettlemoyer2009Learning} or only attempted to learn at very small scales~\cite{gauthier2018word}. We address this combinatory explosion with a novel joint parsing and {\it expected execution} mechanism, namely \ckyee, which extends the classic CKY chart parsing algorithm. 
It performs local marginalization of distributions over sub-programs to make the search process tractable.

In sum, our paper makes three specific contributions.  First, we present the neuro-symbolic \model model that learns grounded lexical representations without requiring annotations for the concepts to be learned or partial word meanings; it automatically recovers underlying concepts in the target domain from language and experience with their groundings. Second, we introduce a novel expected execution mechanism for parsing in model training, to facilitate search in the compositional grammar-based space of meanings. Third, through systematic evaluation on two benchmarks, visual reasoning in CLEVR~\cite{Johnson2017CLEVR} and language-driven navigation in SCAN~\cite{Lake2018Generalization}, we show that the lexicalist design of \model enables learning with strong data efficiency and compositional generalization to novel linguistic constructions and deeper linguistic structures.

\vspace{-1em}
\section{Grammar-Based Grounded Lexicon Learning}
\vspace{-1em}
Our framework, \modelfull (\model) learns grounded lexicons from cross-modal data, such as paired images and texts. Throughout this section, we will be using the visual reasoning task, specifically visual question answering (VQA) as the example, but the idea itself can be applied to other tasks and domains, such as image captioning and language-driven navigation.

\begin{figure*}
    \centering
    \includegraphics[width=\textwidth]{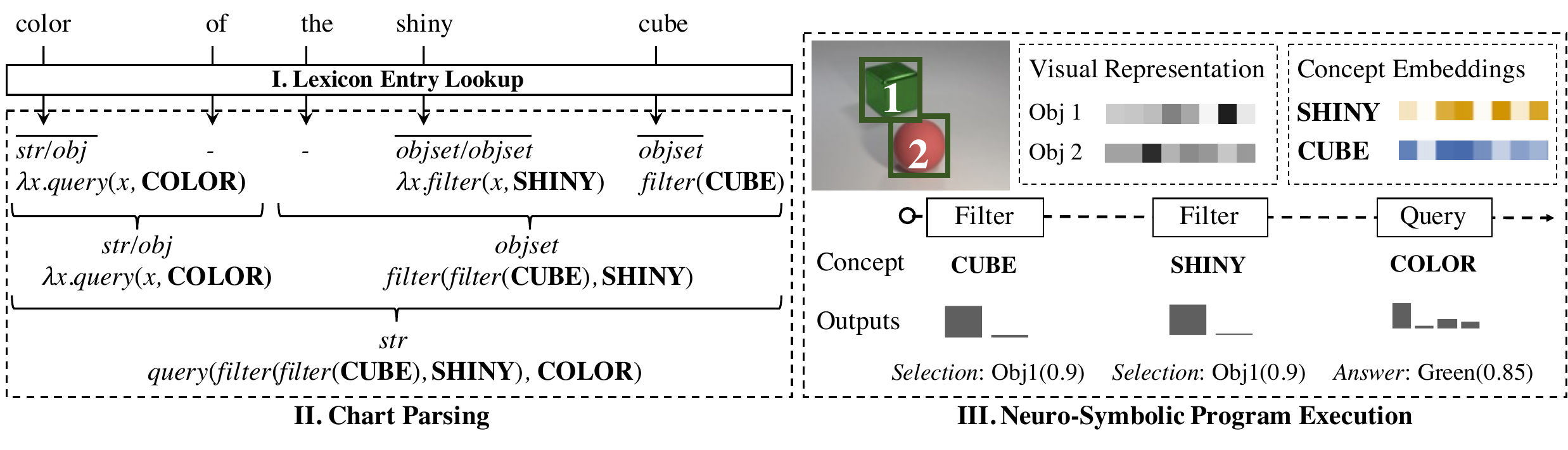}
    \vspace{-2em}
    \caption{\model parses the input sentence into an executable neuro-symbolic program by first (I) lookup the lexicon entry associated with each word, followed by (II) computes the most probable parsing tree and the corresponding tree with a chart parsing algorithm. The derived program can be grounded and executed on an image with a neuro-symbolic reasoning process~\cite{Mao2019NeuroSymbolic} (III).}
    \label{fig:model}
    \vspace{-1em}
\end{figure*}

\model learns from a collection of VQA data tuples, containing an image, a question, and an answer to the question. In \model, each word type $w$ is associated with one or multiple lexical entries, comprised of their syntactic types and semantic programs. Given the input question, \model first looks up the lexicon entries associated with each individual token in the sentence (\fig{fig:model}I). \model then uses a chart parsing algorithm to to derive the programmatic meaning representation of the entire sentence by recursively composing meanings based on syntax (\fig{fig:model}II). To answer the question, we execute the program on the image representation (\fig{fig:model}III). During training, we compare the answer derived from the model with the groundtruth answer to form the supervision for the entire system. No additional supervision, such as lexicon entries for certain words or concept labels, is needed. 

\vspace{-0.5em}
\subsection{Grounded Lexicon}
\vspace{-0.5em}

\begin{wrapfigure}{r}{0.45\textwidth}
    \vspace{-1.7em}
    \centering
    \includegraphics[width=0.45\textwidth]{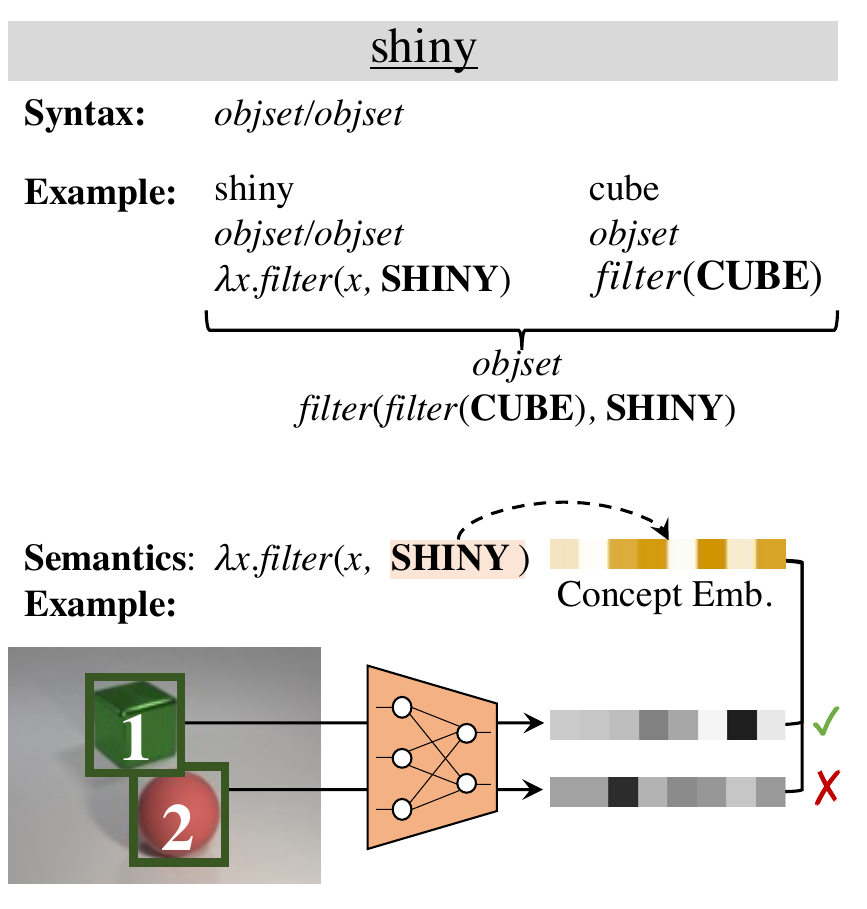}
    \caption{Each word is associated with a grounded lexicon, comprised of its syntactic type and a neuro-symbolic semantic program.}
    \vspace{-2em}
    \label{fig:entry}
\end{wrapfigure}
 
At a high-level, \model follows the combinatory categorical grammar~\cite[CCG; ][]{steedman1996surface} formalism to maintain lexicon entries and parse sentences. Illustrated in \fig{fig:entry}, Each word $w$ (\eg, {\it shiny}) is associated with one or multiple entries. Each entry $e_w^{(i)}$ is a tuple comprised of a syntax type $\textit{syn}_w^{(i)}$ (\eg, {\it objset/objset}), and a semantic meaning form $\textit{sem}_w^{(i)}$ (\eg, $\lambda x.\textit{filter}(x, \textbf{SHINY})$). $\textit{sem}_w^{(i)}$ is a symbolic program represented in a typed domain-specific language (DSL) and can be executed on the input image. Some programs contain concepts (in this case, \textbf{SHINY}) that can be visually grounded.

\myparagraph{Typed domain specific language.}
\model uses a DSL to represent word meanings. For the visual reasoning domain, we use the CLEVR DSL~\cite{Johnson2017CLEVR}. It contains object-level operations such as selecting all objects having a particular attribute (\eg, the shiny objects) or select all objects having a specific relationship with a certain object (\eg, the objects left of the cube). It also supports functions that respond to user queries, such as counting the number of objects or query a specific attribute (\eg, shape) of an object. The language is typed: most functions takes a set of objects or a single object as their inputs, and produce another set of objects. For example, the operation $\textit{filter}$ has the signature $\textit{filter}(\textit{objset}, \textit{concept}) \rightarrow \textit{objset}$ and returns all objects that have $\textit{concept}$ (\eg, all {\it shiny} objects) in the input set.

\myparagraph{Syntactic types.} There are two types of syntactic types in \model: primitive and complex.\footnote{In some domains we also use conjunctions ({\it CONJ}) in the coordination rule.} The primitive types are defined in the typed domain specific language (\eg, {\it objset}, {\it int}). A complex type, denoted as X/Y or X\textbackslash Y, is a functor type that takes an argument of type Y and returns an object of type X. The direction of the slash indicates word order: for X/Y, the argument Y must appear on the right, whereas in X\textbackslash Y, it must appear on the left. Note that X and Y can themselves be complex types, which allows us to define functor types with multiple arguments, such as (X\textbackslash Y)/Z, or even functors with functor arguments (\eg, (X\textbackslash Y)/(Z/Z)).

In \model, the semantic type of a word meaning (in the DSL) together with a set of directional and ordering settings for its arguments (reflecting how the word and its arguments should be linearized in text) uniquely determines the word's syntactic type. For example, the syntactic type for word {\it shiny} is {\it objset/objset}. It first states that {\it shiny} acts as a function in meaning composition, which takes a subprogram that outputs a set of objects (\eg, $\textit{filter}(\textbf{CUBE})$) as its argument, and produces anew program whose output is also a set of objects, in this case,  $\textit{filter}(\textit{filter}(\textbf{CUBE}), \textbf{SHINY})$ Second, it states the direction of the argument, which should come from its right.

\myparagraph{Neuro-symbolic programs.}
Some functions in the DSL involves concepts that will be grounded in other modalities, such as the visual appearance of an object and their spatial relationships. Taking the function $\textit{filter}$ as an example, its secondary argument $\textit{concept}$ should be associated with the visual representation of objects. In \model, the meaning of each lexicon entry may involve one more constants (called ``concepts'') that are grounded on other modalities, possibly via deep neural embeddings. In the case of {\it shiny}: $\lambda x.\textit{filter}(x, \textbf{SHINY})$. The concept $\textbf{SHINY}$ is associated with a vector embedding in a joint visual-semantic embedding space, following \citet{kiros2014unifying}. During program execution, we will be comparing the embedding of concept $\textbf{SHINY}$ with object embeddings extracted from the input image, to filter out all {\it shiny} objects.

\myparagraph{Lexicon learning.} \model learns lexicon entries in the following three steps. (i) First, we enumerate all possible semantic meaning programs derived from the DSL. For example, in the visual reasoning domain, a candidate program is $\lambda x.\textit{filter}(x, \textbf{?})$, where $\textbf{?}$ denotes a concept argument. When we try to associate this lexicon entry to the word {\it shiny}, the program is instantiated as $\lambda x.\textit{filter}(x, \textbf{SHINY})$, where \textbf{SHINY} is a new concept associated with a vector embedding. Typically, we set a maximum number of arguments for each program and constrain its depth. We explain how we set these hyperparameters for different domains in the supplementary material. (ii) Next, for programs that have a primitive type, we use 
its semantic type as the syntactic type (\eg, {\it objset}). For programs that are functions with arguments, we enumerate possible argument ordering of the arguments. For example, the program $\lambda x.\textit{filter}(x, \textbf{SHINY})$ has two candidate syntactic types: {\it objset/objset} (the argument is on its right in language) and {\it objset\textbackslash objset} (the argument is on its left). (iii) Finally, we associate each candidate lexicon entry with a learnable scalar weight $\tau(\cdot)$. It is typical for a single word having tens or hundreds of candidate entries, and we optimize these lexicon entry weights in the training process. In practice, we assume no lexical ambiguity, \ie, {\it each word type has only one lexical entry}.
Thus, the ambiguity of parsing only comes from different syntactic derivation orders for the same lexical entries. This also allows us to prune lexicon entries that do not lead to successful derivations during training.

\vspace{-0.5em}
\subsection{Program Execution}
\vspace{-0.5em}

Any fully grounded programs (\ie, programs without unbound arguments) can be executed based on the image representation. We implement the Neuro-Symbolic Concept Learner~\cite[NS-CL;][]{Mao2019NeuroSymbolic} as our differentiable program executor, which consists of a collection of deterministic functional modules to realize the operations in the DSL. NS-CL represents execution results in a ``soft'' manner: in the visual reasoning domain, a set of objects is represented as a vector mask $m$ of length $N$, where $N$ is the number of objects in the scene. Each element, $m_i$ can be interpreted as the probability that object $i$ is in the set. For example, the operation $\lambda x.\textit{filter}(x, \textbf{SHINY})$ receives an input mask $m$ and produces a mask $m'$ that selects all shiny objects in the input set. The computation has two steps: (i) compare the vector embedding of concept $\textbf{SHINY}$ with all objects in the scene to obtain a mask $m^{(\textbf{SHINY})}$, denoting the probability of each object being {\it shiny}; (ii) compute the element-wise multiplication $m' = m \odot m^{\textbf{SHINY}}$, which can be further used as the input to other functions. In NS-CL, the execution result of any program is fully differentiable \wrt the input image representation and concept embeddings (\eg, $\textbf{SHINY}$).

\begin{figure*}[tp]
\centering
\begin{minipage}[t]{0.64\textwidth}
\vspace{0pt}
\begin{algorithm}[H] 
\caption{The CKY-E$^2$ algorithm.}
\label{alg:ckyee}
\begin{algorithmic}[1]
\small
\Require{$w_i$: the input sentence; $L$: sentence length; $e_i^j$: the $j$-th lexicon entry associated with word $w_i$; $\tau(e_i^j)$: lexicon weights.} 
\Ensure{$\mathit{exe}_k$ the execution result of the all possible derivations and their weights $\tau(\mathit{exe}_k)$.}
\For{$\mathit{i} \gets 0$ to $L - 1$}
    \State Initialize $\mathit{dp}[\mathit{i}, \mathit{i+1}]$ with lexicon entries $e_i^{*}$ and weights $\tau(e_i^*)$
\EndFor
\For{$\mathit{length} \gets 1$ to $L$}
  \For{$\mathit{left} \gets 0$ to $L - \mathit{length}$}
    \State $\mathit{right} \gets \mathit{left} + \mathit{length}$
    \State $\mathit{dp}[\mathit{left}, \mathit{right}] \gets $ empty list
    \For{$\mathit{k} \gets \mathit{left} + 1$ to $\mathit{right} - 1$}
        \State Try to combine nodes in $\mathit{dp}[\mathit{left}, \mathit{k}]$ and $\mathit{dp}[\mathit{k}, \mathit{right}]$
        \State Append successful combination to $\mathit{dp}[\mathit{left}, \mathit{right}]$
    \EndFor
    \State $\textsc{ExpectedExecution}(\mathit{dp}[\mathit{left}, \mathit{right}])$
  \EndFor
\EndFor
\Procedure{ExpectedExecution}{$a$: a list of derivations}
\While{$\exists x, y \in a$ are identical except for subtrees of the same type}
    \State Create $z$ from $x$ and $y$ by computing the expected execution results for non-identical subtrees
    \State $\tau(z) \gets \tau(x) + \tau(y)$
    \State Replace $x$ and $y$ in $a$ with $z$
\EndWhile
\EndProcedure
\end{algorithmic}
\end{algorithm}
\end{minipage}
\hfill
\begin{minipage}[t]{0.33\textwidth}
\vspace{10pt}
\includegraphics[width=1\textwidth]{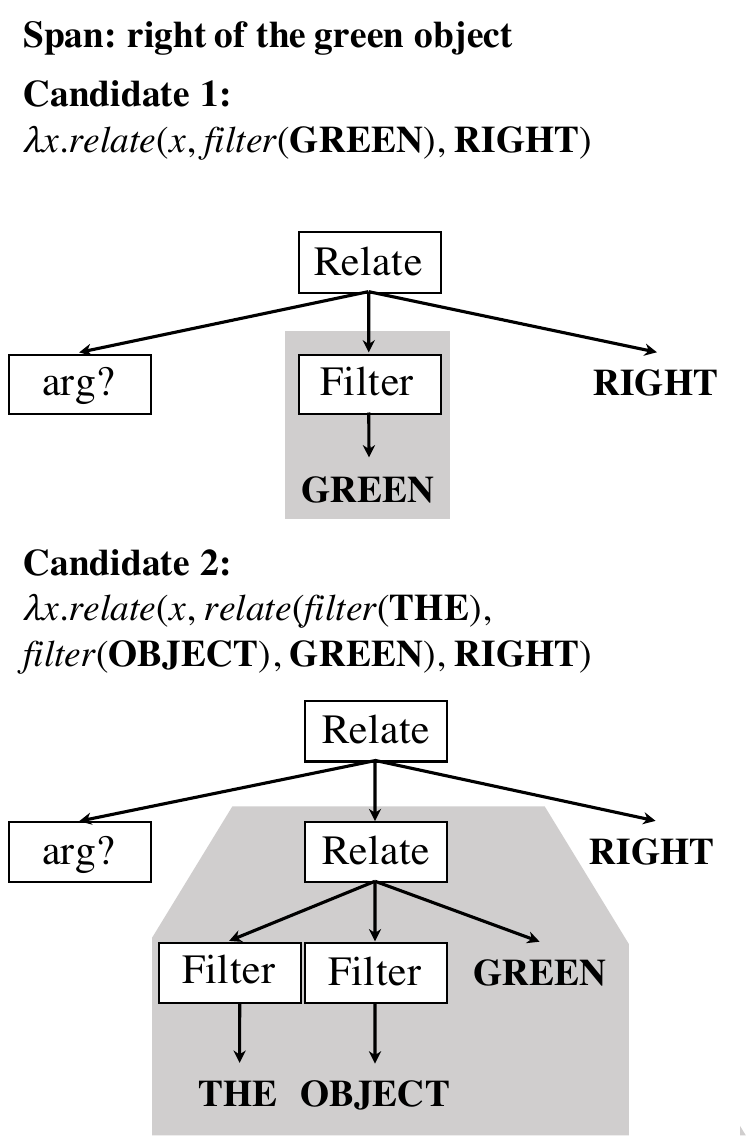}
\captionof{figure}{An illustrative example of two semantic programs that can be merged by computing the expected execution results of two subtrees (highlighted in gray). Both subtrees outputs a vector of scores indicating the objects being selected.}
\label{fig:subtree}
\end{minipage}
\vspace{-2.5em}
\end{figure*}
 
\vspace{-0.5em}
\subsection{Joint Chart Parsing and Expected Execution (\ckyee)}
\vspace{-0.5em}
\model extends a standard dynamic programming algorithm for chart parsing (\ie, the CKY algorithm~\citep{kasami1966efficient,younger1967recognition,cocke1969programming}) to compose sentence meaning from lexical meaning forms, based on syntax. Denote $w_i$ as the input word sequence. $e_{i}^{j}$ the $j$-th lexicon entry associated with word $w_i$, and $\tau(e_{i}^{j})$ the corresponding weight. Consider all possible derivation of the question $\{\textit{derivation}_k\}$, $k=1,2,\dots$. We define the following context-free probability distribution of derivations as:
\begin{center}
\vspace{-1.5em}
\[ p(\textit{derivation}_k) \propto \exp \left( \sum_{e \in \textit{derivation}_k} \tau(e) \right). \]
\vspace{-1.5em}
\end{center}
That is, the probability is exponentially proportional to the total weights $\tau(e)$ of all lexicon entries $e \in \textit{derivation}_k$ used by the specific derivation.

A straightforward implementation to support joint learning of lexicon weights $\tau$ and neural modules (\eg, $\textit{filter}(x, \textbf{SHINY})$), is to simply execute all possible derivations on the input image, and compare the answer with the groundtruth. However, the number of possible derivations grows exponentially as the question length, making such computation intractable. For example, in SCAN~\cite{Lake2018Generalization}, each word has 178 candidate lexicons, and the number of lexicon combination of a sentence with 5 words will be $178^5 \approx 10^{11}$. To address this issue, we introduce the idea of expected execution, which essentially computes the ``expected'' execution result of all possible derivations. We further accelerate this process by taking local marginalization.

Our \ckyee algorithm is illustrated in \alg{alg:ckyee}. It processes all spans $[\textit{left}, \textit{right})$ sequentially ordered by their length. The composition for derivations of $[\textit{left}, \textit{right})$ has two stages. First, it enumerates possible split point $k$ and tries to combine the derivation of $[\textit{left}, k)$ and $[k, \textit{right})$. This step is identical to the standard CKY parsing algorithm. Next, if there are two derivations $x$ and $y$ of span $[i, j)$, whose program structures are identical except for subtrees that can be partially evaluated (\ie, does not contain any unbounded arguments), we will compress these two derivations into one, by marginalizing the execution result for that subtree. 

See the example from \fig{fig:subtree}. Two programs have the identical structure, except for the second argument to the outer-most {\it relate} operation. However, these sub-trees, highlighted in gray, can be partially evaluated on the input image, and both of them output a vector of scores indicating the objects being selected. Denote $\tau_1$ and $\tau_2$ as the weight associated with two derivations, and $v_1$ and $v_2$ the partial evaluation results (vectors) for two subtrees. We will replace these two candidate meaning form with $z$:
\[ z \coloneqq \lambda x.\textit{relate}(x, v', \textbf{RIGHT}),\;\;\text{where~} v'\coloneqq \frac{\exp(\tau_1) v_1 + \exp(\tau_2) v_2}{\exp(\tau_1) + \exp(\tau_2)}, \tau(z) \coloneqq \tau_1 + \tau_2.  \]
We provide additional running examples of the algorithm in the supplementary material.

\myparagraph{Complexity.} Intuitively, once we have determined the semantics of a constituents in the question, the actual concrete meaning form of the derivation does not matter for future program execution, if the meaning form can already be partially evaluated on the input image. This joint parsing and expected execution procedure significantly reduces the exponential space of possible parsing to a polynomial space \wrt the number of possible program layouts that can not be partially evaluated, which, in practice, is small.
The complexity of CKY-E2 is polynomial with respect to the length $L$ of the sentence, and $M$ the number of candidate lexicon entries. More specifically, $O(L^3 M)$, where $O(L^3)$ comes from the chart parsing algorithm, and the number of derivations after the expected execution procedure is $O(M)$. This result is obtained by viewing the maximum arity for functor types being a constant (\eg, 2). Intuitively, for each span, all possible derivations associated with this span can be grouped into 4 categories: derivations of a primitive type, derivations of a 1-ary functor type, derivations of a 2-ary functor type, and derivations of a 2-ary functor type, with one argument binded. All these numbers grow linearly \wrt $M$. For detailed analysis please refer to our supplementary material.

\myparagraph{Correctness.} 
One can theoretically prove that, if all operations in the program layout are commutative with the expectation operator, \ie, if $\mathbb{E}\left([f\left(x\right)\right] = f\left(\mathbb{E}\left[x\right]\right)$, our \ckyee produces exact computation of the expected execution result. These operations include, tensor addition, multiplication (if tensors are independent), and concatenation, which cover most of the computation we will do in neuro-symbolic program execution. For example, for {\it filter}, taking the expectation over different inputs before doing the filtering is the same as taking the expectation over the filter results of different inputs. However, there are operations such as quantifiers whose semantics are not commutative with the expectation operator. In practice, it is possible to still use the expected expectation framework to approximate. We leave the application of other approximated inference techniques as future work. We provide proofs and its connections with other formalisms in the supplementary material.

\vspace{-0.5em}
\subsection{Learning}
\vspace{-0.5em}
Our model, \model, can be trained in an end-to-end manner, by looking at images and reading paired questions and answers. We denote $\ell$ as a loss function that compares the output of a program execution (\eg, a probability distribution over possible answers) and the groundtruth. More precisely, given all possible derivations $\textit{derivation}_k$, the image representation $I$, the answer $A$, and the executor $\mathcal{E}( \cdot, I )$, we optimize all parameters by minimizing the loss $\mathcal L$:
\begin{equation*}
    \mathcal L = \sum_{k} \left( p(\textit{derivation}_k) \cdot \ell \left( \mathcal{E}( \textit{derivation}_k, I ) , A \right) \right).
\end{equation*}
In practice, we use gradient-based optimization for both the neural network weights in concept grounding modules and the lexicon weights $\tau$.
\vspace{-0.5em}
\section{Experiment}
\vspace{-0.5em}
\label{sec:expr}

We evaluate \model on two domains: visual reasoning in CLEVR~\cite{Johnson2017CLEVR} and language-driven navigation in SCAN~\cite{Lake2018Generalization}. Beyond the grounding accuracy, we also evaluate the compositional generalizability and data efficiency, comparing \model with end-to-end neural models and modular neural networks.

\vspace{-0.5em}
\subsection{Visual Reasoning}
\vspace{-0.5em}
\begin{table}[t]
    \centering\small
    \setlength{\tabcolsep}{3pt}
    \begin{tabular}{l cc c cc c ccc c}
    \toprule
         \bf \multirow{2}{*}{\vspace{-5pt}Model} & \bf \multirow{2}{*}{\vspace{-5pt}Prog?} & \bf \multirow{2}{*}{\vspace{-5pt}Concept?} &&  \multicolumn{2}{c}{\bf Standard} && \multicolumn{3}{c}{\bf Compositional Generalization} & \bf Depth \\
         \cmidrule{5-6}  \cmidrule {8-10} 
                                                   &&&& 10\%   & 100\% && \it ~~~purple~~~ & \it ~~~right of~~~ & \it ~~~count~~~ &  \\
    \midrule
         MAC \cite{Hudson2018Compositional}        & N/N & N/N && 85.39 & 98.61  && 97.14 & 90.85 & 54.87  & 77.40  \\ \midrule
         TbD-Net \cite{Mascharka2018Transparency}  & Y/Y & N/N && 44.52 & 98.04  && 89.57 & 49.92 & 63.37  & 53.13\\
         NS-VQA \cite{Yi2018Neuro}                 & Y/Y & Y/Y && \textbf{98.57} & 98.57  && 95.52 & \textbf{99.80} & 81.81  & 50.45 \\
         NS-CL \cite{Mao2019NeuroSymbolic}         & Y/N & Y/N && 98.51 & \textbf{98.91}  && \textbf{98.02} & 99.01 & 18.88  & 81.60 \\
         \midrule
         \model (ours)                             & Y/N & Y/N && 98.11 & 98.25  && 97.82 & 98.59 & \textbf{96.76}  & \textbf{98.49} \\
    \bottomrule \\
    \end{tabular}
    \vspace{-5pt}
    \caption{Accuracy on the CLEVR dataset. Our model achieves a comparable results with state-of-the-art approaches on the standard training-testing split. It significantly outperforms all baselines on generalization to novel word compositions and to sentences with deeper structures. The best number in each column is bolded. The second column indicates whether the model uses program-based representation of question meaning and whether it needs program annotation for training questions. The third column indicates whether the model explicitly models individual concepts and whether it needs concept annotation for objects during training.}
    \label{tab:clevr}
    \vspace{-3em}
\end{table} We first evaluate \model on the visual reasoning tasks in the CLEVR domain~\citep{Johnson2017CLEVR}, where the task is to reason and answer questions about images. In our study, we use a subset of CLEVR dataset, which does not include sentences that involve coreference resolution, and words with multiple meanings in different contexts. We add additional information on how we filter the dataset in the supplementary.

\myparagraph{Setup.} Instead of using manually defined heuristics for curriculum learning or self-paced learning as in previous works~\citep{Mao2019NeuroSymbolic,Li2020Competence}, we employ a curriculum learning setup that is simply based on sentence length: we gradually add longer sentences into the training set. This helps the model to learn basic words from very short sentences (6 words), and use the acquired lexicon to facilitate learning longer sentences (20 words). Since CLEVR does not provide test set annotations, for all models, we held out 10\% of the training data for model development and test them on the CLEVR validation split.

\myparagraph{Baselines.} We compare \model with 4 baselines. (1) MAC~\cite{Hudson2018Compositional} is an end-to-end approach based on attention. (2) TbD-Net~\cite{Mascharka2018Transparency} uses a pre-trained semantic parser to parse the question into a symbolic program, and executes the program with a neural module network~\cite{andreas2016neural}. (3) Similarly, NS-VQA~\cite{Yi2018Neuro} also parses the question into a symbolic program. It also extracts an abstract scene representation with pre-trained neural recognition models~\cite{He2017Mask}. It executes the program based on the abstract scene representation. Both of the approaches require additional supervision for training the semantic parser, and NS-VQA requires additional annotation for training the visual recognition model. (4) NS-CL~\cite{Mao2019NeuroSymbolic} jointly learns a neural semantic parser and concept embeddings by looking at images and reading paired questions and answers. It requires the annotation for all concepts in the domain (\eg, colors and shapes). In contrast, \model can {\it automatically} discover visual concepts from texts.

\myparagraph{Results.}
\tbl{tab:clevr} summarizes the results. We consider any model that performs in the 95–100 range to have more or less solved the task. Small differences in numeric scores in this range, such as the fact that NS-CL outperforms our model on the ``purple'' generalization task by 0.2\%, are less important than the fact that our model far outperforms all competitors on ``count'' compositional generalization and the ``depth'' generalization task, both of which all competitor models are far from solving.

We first compare different models on the {\bf standard} training-testing split. We train different models with either 10\% or 100\% of the training data and evaluate them on the validation set. Our model achieves a comparable performance in terms of its accuracy and data efficiency.

Next, we systematically build three {\bf compositional generalization} test splits: {\it purple}, {\it right of}, and {\it count}. The detailed setups and examples for these splits are provided in the supplementary. Essentially, we remove ~90\% of the sentences containing the word {\it purple}, the phrase {\it right}, and {\it counting operations}, such as {\it how many ...?} and {\it what number of ...?}. We only keep sentences up to a certain length (6 for purple, 11 for right, and 8 for count). We make sure that each use case of these words appear in training questions. After training, we test these models on the validation set with questions containing these words. Overall, our model \model outperforms all baselines on all three generalization splits. In particular, it significantly outperforms other methods on the {\it count} split. The {\it count} split is hard for other method because it requires model to generalize to sentences with deeper structures, for example, from ``{\it how many red objects are there?}'' to ``{\it how many red objects are right of the cube?}'' Note that, during training, all models have seen example use of similar structures such as ``{\it what's the shape of the red object}'' and ``{\it what's the shape of the red object right of the cube?}''

Finally, we test generalization to sentences with deeper structures ({\bf depth}). Specifically, we define the ``hop number'' of a question as the number of intermediate objects being referred to in order to locate the target object. For example, the ``hop number'' of the question ``{{\it how many red objects are right of the cube?}}'' is 1. We train different models on 0-hop and 1-hop questions and test them on 2-hop questions. Our model strongly outperforms all baselines.

The results on the {\bf compositional generalization} and {\bf depth} splits yield two conclusions. First, disentangling grounded concept learning (associating words onto visual appearances) and reasoning (e.g., filtering or counting subsets of objects in a given scene) improves data efficiency and generalization. On CLEVR, neuro-symbolic approaches that separately identify concepts and perform explicit reasoning (NS-VQA, NS-CL and \model) consistently generalize better than approaches that do not (MAC, TbD). The comparison between TbD and NS-VQA is informative: TbD fails on the “right of” task even in the case where the semantic parser is providing correct programs, while NS-VQA, which uses the same parser but explicitly represents compositional symbolic concepts for reasoning, succeeds in this task. Crucially, of the three neuro-symbolic methods, \model achieves strong performance with less domain-specific knowledge than other methods: NS-VQA needs groundtruth programs; NS-CL needs the concept vocabulary; \model requires neither. Second, our model is the only one to perform well on the hardest ``out-of-sample'' generalization tests: holding out ``count'' and generalizing to deeper embeddings. The other, easier generalization tests all have close neighbors in the training set, differing by just one word. In contrast, the length, depth and ``count'' tests require generalizing to sentences that differ in multiple words from any training example. They appear to require -- or at least benefit especially well from -- \model’s lexical-grammatical approach to capturing meaning of complex utterances, with explicit constituent-level (as opposed to simply word-level) composition. We also provide in-depth analysis for the behavior of different semantic parsing models in the supplementary material.

\vspace{-1em}
\subsection{Language-driven Navigation}
\vspace{-1em}
\label{sec:scan}
\begin{table}[t]
    \centering\small
    \setlength{\tabcolsep}{6pt}
    \begin{tabular}{l cc c     cc c}
    \toprule
         \bf \multirow{2}{*}{\vspace{-5pt}Model} &  \multicolumn{2}{c}{\bf Simple} && \multicolumn{2}{c}{\bf Compositional Generalization} & \bf Length \\
         \cmidrule{2-3}  \cmidrule {5-6} 
                   & 10\%  & 100\% &&  \it jump & \it around right \\
    \midrule
         seq2seq \citep{Sutskever2014Sequence}      & 0.93\stderr{0.05} & 0.99\stderr{0.01}          %
         && 0.00\stderr{0.00}$^\dagger$ & 0.00\stderr{0.00}$^\dagger$ & 0.15\stderr{0.02} \\
         Transformer \citep{Vaswani2017Attention}   & 0.71\stderr{0.24} & 0.78\stderr{0.11}%
         && 0.00\stderr{0.00}~~ & 0.10\stderr{0.08}~~ & 0.02\stderr{0.01} \\
         GECA \citep{andreas-2020-good}              & 0.99\stderr{0.00} &   0.98\stderr{0.01}                  %
         && 0.87\stderr{0.05}$^\dagger$ & 0.82\stderr{0.11}$^\dagger$ & 0.15\stderr{0.02} \\
         WordDrop \citep{guo-etal-2020-sequence}$^*$     & 0.56\stderr{0.02}                & 0.62\stderr{0.02}       %
         && 0.52\stderr{0.02}~~    & 0.70\stderr{0.06}~~ & 0.18\stderr{0.01} \\
         SwitchOut \citep{wang2018switchout}$^*$    & 0.99\stderr{0.01}                & 0.99\stderr{0.01}                         %
         && 0.98\stderr{0.02}~~    & 0.97\stderr{0.02}~~ & 0.17\stderr{0.02} \\ 
         SeqMix \citep{guo-etal-2020-sequence}$^*$       & --                & --                         %
         && 0.98$^\ddagger$~~~~~~~~~    & 0.89$^\ddagger$~~~~~~~~~ & -- \\
         recomb-2 \citep{akyurek-etal-2020-learning} & --                & --                         %
         && 0.88\stderr{0.07}$^\dagger$ & 0.82\stderr{0.08}$^\dagger$ & -- \\
         \model (ours)                               & \textbf{1.00}\stderr{0.00}  & \textbf{1.00}\stderr{0.00} & & \textbf{1.00}\stderr{0.00}~~~ & \textbf{1.00}\stderr{0.00}~~~& \textbf{1.00}\stderr{0.00}\\
    \bottomrule \\
    \end{tabular}
    \vspace{-5pt}
    \caption{Accuracy on the SCAN dataset, averaged across 10 valid runs when applicable, $\pm$ denotes standard deviation. The best number in each column is bolded. $\dagger$: results taken from \citep{akyurek-etal-2020-learning}; $\ddagger$: results taken from \citep{guo-etal-2020-sequence}. Both paper have only presented results on the compositional generalization split. $*$: applied after GECA. The results for GECA are based on the released implementation by the authors. All the models are selected with respect to the accuracy on the training set. }\vspace{-10pt}
    \label{tab:scan}
    \vspace{-1.5em}
\end{table} The second domain we consider is language-driven navigation. We evaluate models on the SCAN dataset~\cite{Lake2018Generalization}: a collection of sentence and navigational action sequence pairs. There are 6 primitive actions: {\it jump}, {\it look}, {\it walk}, {\it run}, {\it lturn}, and {\it rturn}, where an instruction {\it turn left twice and run} will be translated to {\it lturn lturn run}. All instructions are generated from a finite context-free grammar, so that we can systematically construct train-test splits for different types of compositional generalizations.

\myparagraph{Setup.} We use a string-editing domain-specific language (DSL) for modeling the meaning of words in the SCAN dataset, of which the details can be found in the supplementary material. At a high level, the model supports three primitive operations: constructing a new constant string (consisting of primitive operations), concatenating two strings, and repeating the input string for a number of times.

For \model, we generate candidate lexicons by enumerating functions in the string-editing DSL with up to 2 arguments and the function body has a maximum depth of 3. We also allow at most one of the argument being functor-typed, for example, {\it V\textbackslash V/(V\textbackslash V)}. To handle parsing ambiguities, we use two primitive syntax types $S$ and $V$, while both of them are associated with the semantic type of {\it string}. In total, we have 178 candidate lexicon entries for each word.

\myparagraph{Baselines.} We compare \model to seven baselines. (1) Seq2seq \citep{Sutskever2014Sequence} trains an LSTM-based encoder-decoder model. We follow the hyperparameter setups of \citep{Lake2018Generalization}. (2) Transformer~\citep{Vaswani2017Attention} is a 4-head Transformer-based autoregressive seq2seq model. We tuned the hidden size (\ie, the dimension of intermediate token representations) within \{100, 200, 400\}, as well as the number of layers (for both the encoder and the decoder) from \{2, 4, 8\}. Other methods are based on different data augmentation schemes for training a LSTM seq2seq model. Specifically, (3) GECA augments the original training splits using heuristic span recombination rules; (4) WordDrop~\citep{guo-etal-2020-sequence} performs random dropout for input sequence (while keeping the same label); (5) similarly, SwitchOut \citep{wang2018switchout} randomly replaces an input token with a random token from the vocabulary; (6) SeqMix~\citep{guo-etal-2020-sequence} uses soft augmentation techniques following \citep{zhang2017mixup}, which composes an ``weighted average'' of different input sequences; (7) recomb-2~\citep{akyurek-etal-2020-learning} learns recombination and resampling rules for augmentation.

\myparagraph{Results.} We compare different models on three train-test splits. In {\bf Simple}, the training and test instructions are drawn from the same distribution. We compare the data efficiency of various models by using either 10\% or 100\% of the training data, and test them on the same test split. While all models can achieve a nearly-perfect accuracy with 100\% training data, our model \model shows advantage with only a small amount of data. Next, in {\bf Compositional}, we have held out the sentences containing certain phrases, such as {\it jump} and {\it around right}. For these held-out phrases, only valid non-contextual examples containing them (\ie, \textit{jump} in isolation and no example for \textit{around right}) are available during training. During test, algorithms need to make systematical generalization of these phrases in novel contexts. Finally, in {\bf Length}, all training examples have the action length less than or equal to 22, while that of a test example is up to 48. Our model consistently reach perfect performance in all considered settings, even on the cross-length generalization task where GECA does not help improve performance.
These results are consistent with the conclusions we derived on the CLEVR dataset. Specifically, data-augmentation techniques for SCAN can solve simple generalization tests (\eg, {\it jump}, where tests all have close neighbors in the training set, differing by just one word) but not the hard ones (\eg, {\it length}, where test sentences can different in multiple words from any training examples).

\myparagraph{Cases study.}
\model is expressive enough to achieve perfect accuracy on the SCAN dataset: there exists a set of lexicon entries which matches the groundtruth in SCAN. However, our learning algorithm does not always converge on the correct lexicon, but when it fails, the failure can be identified based on training-set accuracy. So, we perform model selection based on the training accuracy for \model: after a sufficient number of epochs, if the model hasn’t reached perfect accuracy (100\%), we re-initialize the weights and train the model again. Our results show that, among 100 times of training, the model reaches 100\% accuracy 74\% of the time. For runs that don’t have 100\% accuracy, the average performance is 0.94.

Since \model directly learns human-interpretable lexicon entries associated with each individual words, we can further inspect the failure cases made by it when the training accuracy does not converge to 0. We find that the most significant failure mode is the word {\it and} (\eg, {\it jump and run}) and {\it after} (\eg, {\it jump after run}). Both of them are treated as connectives in SCAN. Sometimes \model fails to pick the syntax type {\it S\textbackslash V/V} over the type {\it V\textbackslash V/V}. The entry {\it V\textbackslash V/V} will succeed in parsing most cases (\eg, {\it jump and run}), except that it will introduce ambiguous parsing for sentences such as ``{\it jump and run twice}'': {\it jump and \underline{run twice}} {\it vs.} {\it \underline{jump and run} twice}. Based on the definition of the SCAN, only the first derivation is valid. In contrast, using {\it S\textbackslash V/V} resolves this ambiguity. Depending on the weight initialization and the example presentation order, \model sometimes get stuck at the local optima of {\it V\textbackslash V/V}. However, we can easily identify this by the training accuracies---\model is able to reach perfect performance on all considered splits by simply retraining with another random seed, therefore, we only select those with 100\% training accuracy as valid models. 

\vspace{-0.5em}
\section{Related Work}
\vspace{-0.5em}

\myparagraph{Lexicalist theories.}
The lexicalist theories of syntax~\citep{pollard-sag:1994,steedman:2000,bresnan-etal:2016-lexical} propose that 1) the key syntactic principles by which words and phrases combine are extremely simple and general, and 2) nearly all of the complexity in syntax can be attributed to rich and detailed lexical entries for the words in the language.
For example, whereas the relationship between the active and passive voice, \eg, ``Kim saw a balloon'' versus ``A balloon was seen by Kim'', was treated in pre-lexicalist theories as a special syntactic rule converting between the sentences, in lexicalist theories this relationship derives simply from the knowledge that the passive participle for the verb ``see'' is ``seen,'' which interacts with knowledge of other words to make both the active and passive forms of the sentence possible.
In lexicalist theories, the problem for the language learner thus becomes a problem of learning the words in the language, not a problem of learning numerous abstract rule schemas. The combinatory categorial grammar~\citep[CCG; ][]{steedman1996surface} framework we use is a well-established example of a lexicalist theory: there is a universal inventory of just three combinatory rules (\fig{fig:teaser}a), but those rules can only be applied once richly specified lexical entries are learned for the words in a sentence. We believe that this lexicalist-theory approach is a particularly good fit to the problem of grounded language learning: the visual context provides clues to the word’s meaning, and the word’s grammatical behavior is tied closely to this meaning, making learning efficient.

\myparagraph{Compositional generalization in NLP.} 
Improving the compositional generalization of natrual langauge processing (NLP) systems have drawn great attention in recent years~\citep{baroni2020linguistic}. Most of the recent approaches towards this goal are mostly built on deep learning-based models. There are two representative approaches: building structured neural networks with explicit phrase-based structures or segments~\citep{socher2013recursive,zhu2015long,tai2015improved,Saque2020Multimodal}; and using data augmentation techniques~\citep{andreas-2020-good,guo-etal-2020-sequence,akyurek-etal-2020-learning}. However, these approaches either rely on additional annotation or pretrained models for phrase structure inference or require domain-specific heuristics in data augmentation. In contrast to both approaches, we propose to use combinatory grammar rules to constrain the learning of word meanings and how they compose. 

\myparagraph{Neural latent trees.} \ckyee is in spirit related to recent work using CKY-style modules for inducing latent trees. However, our model is fundamentally different from works on unsupervised constituency parsing~\citep{kim2019compound,shi2021learning} which use the CKY algorithm for inference over scalar span scores and those compute span representation vectors with CKY-style algorithms~\citep[][\textit{inter alia}]{maillard2018latent,drozdov2019unsupervised}. Our key contribution is to introduce the expected execution mechanism, where each span is associated with weighted, compressed programs. Beyond enumerating all possible parsing trees as in~\citep{maillard2018latent}, G2L2 considers all possible programs associated with each span. Our expected execution procedure works for different types (object set, integer, etc.) and even functor types. This makes our approximation exact for linear cases and has polynomial complexity.

\myparagraph{Grammar-based grounded language learning.}
There have also been approaches for learning grammatical structures from grounded texts~\cite{Shi2019Visually,Zhao2020Visually,jin-schuler-2020-grounded,Artzi2013Weakly,Ross2018Grounding,tellex2011understanding}. However, these approaches either rely on pre-defined lexicon entries~\cite{Artzi2013Weakly}, or only focus on inducing syntactic structures such as phrase-structure grammar~\cite{Shi2019Visually}. Different from them, \model jointly learns the syntactic types, semantic programs, and concept grounding, only based on a small set of combinatory grammar rules.

Grammar-based and grounded language learning have also been studied in linguistics, with related work to ours studying on how humans use grammar as constraints in learning meaning~\cite{steedman:2000} and how learning syntactic rules and semantic meanings in language bootstrap each other\cite{abend2017bootstrapping,taylor1988adjectives}. However, most previous computational models have focused only on explaining small-scale lab experiments and do not address grounding in visual perception~\cite{fazly2010probabilistic,gauthier2018word}. In contrast, \model is a neuro-symbolic model that integrates the combinatory categorial grammar formalism~\cite{steedman1996surface} with joint perceptual learning and concept learning, to directly learn meanings from images and texts.

\myparagraph{Neuro-symbolic models for language grounding.} Integrating symbolic structures such as programs and neural networks has shown success in modeling compositional queries in various domains, including image and video reasoning~\cite{Hu2017Learning,Mascharka2018Transparency}, knowledge base query~\cite{Andreas2016Learning}, and robotic planning~\cite{andreas2017modular}. In this paper, we use symbolic domain-specific languages with neural network embeddings for visual reasoning in images and navigation sequence generation, following NS-CL~\cite{Mao2019NeuroSymbolic}. However, in contrasts to using neural network-based semantic parser as in the aforementioned papers, our model \model focuses on learning grammar-based lexicon for compositional generalization in linguistic structures, such as novel word composition.

\vspace{-10pt}
\section{Conclusion and Discussion}
\vspace{-10pt}
\label{sec:conclusion}

In this paper, we have presented \model, a lexicalist approach towards learning compositional and grounded meaning of words. \model builts in a compact but potentially universal set of combinatory grammar rules and learns grounded lexicon entries from a collection of sentences and their grounded meaning, without any human annotated lexicon entries. The lexicon entries represent the semantic type of the word, the ordering settings for its arguments, as well as the grounding of concepts in its semantic program. To facilitate lexicon entry induction in an exponentially-growing space, we introduced \ckyee for joint chart parsing and {\it expected execution}.

Through systematical evaluation on both visual reasoning and language-driven navigation domains, we demonstrate the data efficiency and compositional generalization capability \model, and its general applicability in different domains. The design of \model suggests several research directions. First, in \model we have made strong assumptions on the context-independence of the lexicon entry as well as the application of grammar rules, the handling of linguistic ambiguities and pragmatics needs further exploration~\cite{Frank2012Predicting}. Second, meta-learning models that can leverage learned words to bootstrap the learning of novel words, such as syntactic bootstrapping~\cite{gauthier2018word}, is a meaningful direction. Finally, future work may consider integrating \model with program-synthesis algorithms~\cite{ellis2020dreamcoder} for learning of more generic and complex semantic programs.

\myparagraph{Broader impact.} The ideas and techniques in this paper can be potentially used for building machine systems that can better understand the queries and instructions made by humans.
We hope researchers and developers can build systems for social goods based on our paper.
Meanwhile, we are aware of the ethical issues and concerns that may arise in the actual deployment of such systems, particularly biases in language and their grounding. The strong interpretability of the syntactic types and semantic programs learned by our model can be used in efforts to reduce such biases.

\myparagraph{Acknowledgements.} We thank Wentao Wang for identifying and correcting a mathematical error in our paper. This work is in part supported by ONR MURI N00014-16-1-2007, the Center for Brain, Minds, and Machines (CBMM, funded by NSF STC award CCF-1231216), the MIT Quest for Intelligence, Stanford Institute for Human-Centered AI (HAI), Google, MIT-IBM Watson AI Lab, Samsung GRO, and ADI. Any opinions, findings, and conclusions or recommendations expressed in this material are those of the authors and do not necessarily reflect the views of our sponsors.
 
{
\small
\bibliographystyle{plainnat}
\bibliography{reference,nccg,nscl}
}

\clearpage

\begin{center}
    \LARGE \bf Supplementary Material for\\ Grammar-Based Grounded Lexicon Learning
\end{center}
\vspace{1em}

\appendix
In the supplementary material, we describe the domain specific languages used in our experiments (Section~\ref{sec:1-dsl}), demonstrate how the proposed \ckyee method works by a concrete example (Section~\ref{sec:2-1-example}), show formal properties of \ckyee (Section~\ref{sec:2-2-proof}), present dataset setups and analyze model behaviors (Section~\ref{sec:3-datasets}).
\newcommand{\sigmadot}[2]{\sigma\left( \langle #1, e_{\textbf{#2}} \rangle \right)}

\section{Domain Specific Languages and Neuro-Symbolic Reasoning}
\label{sec:1-dsl}

In this section, we will present and discuss the domain-specific languages (DSLs) we use for two domains: visual reasoning and language-guided navigation. We will further introduce the neuro-symbolic module we have designed for executing programs in these two domains. Overall, each DSL contains a set of types and a set of deterministic modules that have been manually designed for realizing necessary operations in these domains. However, in contrast to realizing them as we do in standard programming languages (with for-loops and if-conditions), we will be using tensor operations (\eg, tensor additions and multiplications) to realize them so that the output of each program is differentiable with respect to all of its inputs.

\subsection{Visual Reasoning DSL}
Our domain-specific language (DSL) for the visual reasoning domain is based on the CLEVR DSL introduced in \cite{Johnson2017CLEVR}, and the neuro-symbolic realization of each functional module is slightly modified from the Neuro-Symbolic Concept Learner~\cite[NS-CL; ][]{Mao2019NeuroSymbolic}. We refer readers to the original papers for a detailed introduction to the DSL and neuro-symbolic program execution. Here we only highlight the key aspects of our language and its neuro-symbolic realization, and discuss the difference between our implementation and the ones in original papers.

Our visual reasoning DSL is a subset of CLEVR, containing 6 types and 8 primitive operations. \tbl{tab:clevr-typesystem} illustrates all 6 types and how they are internally represented in neuro-symbolic execution.

\begin{table}[ht]
    \centering
    \begin{tabular}{lp{0.25\columnwidth}p{0.5\columnwidth}} \toprule
        Type & Note & Representation \\ \midrule
        ObjConcept & Object-level concepts. & An embedding vector. \\ \midrule
        Attribute & Object-level attributes. & A vector of length $K_{\textit{obj}}$, where $K_{\textit{obj}}$ is the number of  \\ \midrule
        RelConcept & Relational concepts. & An embedding vector. \\ \midrule
        ObjectSet & A set of objects in the scene. & A vector $\mathbf{m}$ of length $N$, where $N$ is the number of objects in the scene. Each entry $\mathbf{m}_i$ is a real value in $[0, 1]$, which can be interpreted as the probability that object $i$ is in this set. \\ \midrule
        Integer & An integer. & A single non-negative real value, which can be interpreted as the ``expected'' value of this integer. \\ \midrule
        Bool & A Boolean value. & A single real value in $[0, 1]$, which can be interpreted as the probability that this Boolean value is true. \\ \bottomrule
    \end{tabular}
    \vspace{5pt}
    \caption{The type system of the domain-specific language for visual reasoning.}
    \label{tab:clevr-typesystem}
\end{table}

\begin{table}[tp]
    \centering
    \setlength{\tabcolsep}{3pt}
    \begin{tabular}{p{0.55\columnwidth}p{0.40\columnwidth}} \toprule
        Signature & Note\\ \midrule
        {\it scene}() $\longrightarrow$ ObjectSet & Return all objects in the scene.\\ \midrule
        {\it filter}($\mathbf{a}$: ObjectSet, $c$: ObjConcept) $\longrightarrow$ ObjectSet & Filter out a set of objects having the object-level concept (\eg, red) from the input object set. \\ \midrule
        {\it relate}($\mathbf{a}$: ObjectSet, $\mathbf{b}$: ObjectSet, $c$: RelConcept) $\longrightarrow$ ObjectSet & Filter out a set of objects in set $\mathbf{a}$ that have the relational concept (\eg, left) with the input object $\mathbf{b}$. \\ \midrule
        {\it intersection}($\mathbf{a}$: ObjectSet, $\mathbf{b}$: ObjectSet) $\longrightarrow$ ObjectSet & Return the intersection of set $\mathbf{a}$ and set $\mathbf{b}$. \\ \midrule
        {\it union}($\mathbf{a}$: ObjectSet, $\mathbf{b}$: ObjectSet) $\longrightarrow$ ObjectSet & Return the union of set $\mathbf{a}$ and set  $\mathbf{b}$.\\ \midrule
        {\it query}($\mathbf{a}$: ObjectSet, $c$: Attribute) $\longrightarrow$ ObjConcept & Query the attribute (\eg, color) of the input object $\mathbf{a}$.\\ \midrule
        {\it exist}($\mathbf{a}$: ObjectSet) $\longrightarrow$ Bool & Check if the set is empty.\\ \midrule
        {\it count}($\mathbf{a}$: ObjectSet) $\longrightarrow$ Integer & Count the number of objects in the input set.\\ \bottomrule
    \end{tabular}
    \vspace{5pt}
    \caption{All operations in the domain-specific language for visual reasoning.}
    \label{tab:clevr-dsl}
\end{table}

\tbl{tab:clevr-dsl} further shows all operations in the DSL. There are two main differences between the DSL used by \model and the original CLEVR DSL. First, we have removed the \textit{unique} operation, whose semantic meaning was to return the single object in a set of objects. For example, it can be used to represent the meaning of word ``{\it the}'' in ``{\it \underline{the} red object}'', in which the semantic program of ``{\it red object}'' yields a set of red objects and the semantic program of ``{\it the}'' selects the unique object in that set. However, the meaning of ``{\it the}'' may have slightly different semantic type in different contexts, for example, ``{\it what is \underline{the} color of ...}''. Since this has violated our assumption about each word having only one lexicon entry, we choose to remove this operation to simplify the learning problem. Meanwhile, to handle the ``uniqueness'' of the object being referred to, in our realization of related operations, such as {\it relate} and {\it query}, we will implicitly choose the unique object being referred to, which we will detail in the following paragraphs.

\paragraph{Object-centric scene representation.} In our visual reasoning domain, we have assumed access to a pre-trained object-detector that generates a list of bounding boxes of objects in the scene. In our implementation, following Mao~\etal~\citep{Mao2019NeuroSymbolic}, we use a pre-trained Mask R-CNN~\cite{He2017Mask} to generate bounding boxes for each object proposal. These bounding boxes, paired with the original image, are then sent to a ResNet-34~\citep{He2015Deep} to extract a region-based representation (by RoI Align) and image-based representation, respectively. We concatenate them to form a vector embedding for each object in the image.

\paragraph{Neuro-symbolic realization.} The high-level idea for the program execution is to build a collection of functions that realize the semantics of each operation based on the vector embeddings of objects and concepts. Taking the {\it filter} operation as an example, denote $\mathbf{a}$ as a vector representation of the input set, $o_i$ the object embeddings, and $e_c$ the concept embedding. We compute the vector representation $\mathbf{b}$ of the output set as:
\[ \mathbf{b}_i = \mathbf{a_i} \cdot \sigma\left(\left\langle o_i, e_c \right\rangle\right), \]
where $\sigma$ is the sigmoid function, and $\langle \cdot, \cdot, \rangle$ is the inner product of two vectors. Intuitively, we first compute the inner product between the concept embedding $e_c$ and each object embedding, which gives as a vector of scores of whether object $i$ has concept $c$. Next, we compute the element-wise multiplication between two vectors. 

A key difference between our realization of these operations and the one in Mao~\etal~\cite{Mao2019NeuroSymbolic} is that we use element-wise multiplication to simulate the intersection between two sets, and $1 - (1 - \mathbf{a})(1 - \mathbf{b})$ for union. In contrast, Mao~\etal~\cite{Mao2019NeuroSymbolic} use element-wise min operation for intersection and max for union. Both realizations can be motivated by real-valued logic: product logic \vs G\"odel logic. The main purpose of using products instead of min-max's is to make our realization compatible with our expected execution mechanism, which we will detail in \sectapp{app:ckyee}.

\paragraph{Example.} Here we run a concrete example to illustrate the execution process of a program in the visual reasoning domain. Suppose we have an image containing three objects $o_1$, $o_2$ and $o_3$. We have two additional vector embeddings for concepts \textbf{SHINY} and \textbf{CUBE}. Furthermore, $\sigmadot{o_i}{SHINY} = [0.1, 0.8, 0.9]$, and $\sigmadot{o_i}{CUBE} = [0.8, 0.1, 0.9]$.

Consider the input sentence ``{\it How many shiny cubes are there}''. \tbl{tab:clevr-execution} illustrates a step-by-step execution of the underlying program: $\textit{count}(\textit{filter}(\textit{filter}(\textit{scene}(), \textbf{CUBE}), \textbf{SHINY}))$.

\begin{table}[t]
    \centering
    \begin{tabular}{lll} \toprule
        Program & Type & Value \\ \midrule
        $\textit{scene}()$ & ObjectSet & $[1, 1, 1]$ \\ \midrule
        $\textit{filter}(\textit{scene}(), \textbf{CUBE})$ & ObjectSet & $[0.8, 0.1, 0.9]$ \\ \midrule
    \mycell{$\textit{filter}(\textit{filter}(\textit{scene}(), \textbf{CUBE}), \textbf{SHINY})$} & \mycell{ObjectSet} & \mycell{$[0.08, 0.08, 0.81]$\\=$[0.8, 0.1, 0.9] \odot [0.1, 0.8, 0.9]$} \\ \midrule
        $\textit{count}(\textit{filter}(\textit{filter}(\textit{scene}(), \textbf{CUBE}), \textbf{SHINY}))$ & Integer & $0.97 = \textit{sum}([0.08, 0.08, 0.81])$ \\ \bottomrule
    \end{tabular}
    \vspace{5pt}
    \caption{An illustrative execution trace of the program $\textit{count}(\textit{filter}(\textit{filter}(\textit{scene}(), \textbf{CUBE}), \textbf{SHINY}))$. $\textit{sum}$ denotes the ``reduced sum'' operation of a vector, which returns the summation of all entries in that vector. $\odot$ denotes  element-wise multiplication for two vectors.}
    \label{tab:clevr-execution}
\end{table}

\paragraph{Expected execution.}
In the visual reasoning domain, we have only implemented the expected execution mechanism for subordinate program trees whose type is {\it objset}, although many other types such as {\it integer} and {\it bool} also naturally supports expected execution. This is because, types such as {\it integer} and {\it bool} only appear at the sentence-level, and thus computing the ``expectation'' of such programs does not reduce the overall complexity.

Formally, the expected execution process compresses a list of semantic programs $v_1, v_2, \cdots, v_K$ and their corresponding weights $\tau(v_i)$ into a single semantic program $v^*$ with weight $\tau(v^*)$. Suppose all $v_i$'s have type {\it objset}. We use $\bar{v}_i$ to denote the execution result of these programs. Each of them is a vector of length $N$, where $N$ is the number of objects in the scene. We compute $\bar{v}^*$ and $\tau(v^*)$ as the following:
\begin{eqnarray*}
    \bar{v}^* & = & \frac{1}{\sum_i \exp(\tau(v_i))} \sum_i \left( \exp(\tau(v_i)) \cdot \bar{v}_i \right),\\
    \tau(v^*) & = & \log \sum_i \exp( \tau(v_i) ).
\end{eqnarray*}
Intuitively, we normalize the weights using a softmax function to translate them into a distribution. Then we compute the expectation of the vectors. For more details about the definition and properties of expected execution, please refer to our main text and \sectapp{app:ckyee}.

\paragraph{Candidate lexicons.} Recall that the process of lexicon learning has three stages. First, we generate an extensive collection of candidate semantic programs. Second, we generate candidate lexicon entries for each word by enumerating all possible candidate semantic programs generated in the first step and all possible ordering (linearization in a sentence) of its arguments. Third, we apply our \ckyee and gradient-based optimization to update the weights associated with each lexicon entry.

In our visual reasoning domain, we only consider the following candidate semantic programs and linearizations:
\begin{enumerate}
    \item  Syntactic type: $\textit{objset}$, semantic program: $\textit{scene}()$ (English noun).
    \item  Syntactic type: $\textit{objset}$, semantic program: $\textit{filter}(\textit{scene}(), \textbf{?})$ (English noun).
    \item  Syntactic type: $\textit{objset}/\textit{objset}$, semantic program: $\lambda x.\textit{filter}(x, \textbf{?})$ (English adjective).
    \item  Syntactic type: $\textit{objset}\backslash\textit{objset}/\textit{objset}$, semantic program: $\lambda x.\lambda y.\textit{relate}(x, y, \textbf{?})$ (English preposition I).
    \item  Syntactic type: $\textit{objset}\backslash\textit{objset}/\textit{objset}$, semantic program: $\lambda x.\lambda y.\textit{relate}(y, x, \textbf{?})$ (English preposition II)
    \item  Syntactic type: $\textit{bool}/\textit{objset}$, semantic program: $\lambda x.\textit{exist}(x)$.
    \item  Syntactic type: $\textit{integer}/\textit{objset}$, $\lambda x.\textit{count}(x)$.
    \item  Syntactic type: $\textit{word}/\textit{objset}$, $\lambda x.\textit{query}(x, \textbf{?})$.
    \item  Syntactic type: $\text{CONJ}_{\text{AND}}$, $\lambda f.\lambda g.(\lambda x.\textit{intersect}(f(x), g(x)))$ (generalized conjunction).
    \item  Syntactic type: $\text{CONJ}_{\text{OR}}$, $\lambda x.\lambda y.(\lambda z.\textit{intersect}(z, \textit{union}(x, y)))$ (generalized disjunction).
\end{enumerate}

As we will see later, when we compare the candidate lexicon entries for the visual reasoning domain and the language-driven navigation domain, the visual reasoning domain contains significantly fewer entries than the navigation domain. This is because much of the learning process in this domain is associated with learning the concept embeddings. In the following few paragraphs, we will explain how we instantiate concepts based on these lexicon entry templates and implement generalized conjunction and disjunction in our domain.

First, for each word (more precisely, word type), \eg, {\it shiny}, we will instantiate 10 lexicon entries. For semantic programs that contain unbounded concept arguments (\textbf{?} marks), we will introduce a series word-type-specific concepts. Specifically in this domain, each word type will be associated with 3 concept representations: $\textbf{SHINY}_{\text{ObjConcept}}$, $\textbf{SHINY}_{\text{RelConcept}}$, and $\textbf{SHINY}_{\text{Attribute}}$. Based on \tbl{tab:clevr-dsl}, the first two concepts will be represented as two embedding vectors, and the the third concept will be represented as a vector, indicating which concepts belong to this attribute category. Next, we will instantiate these lexicon entries by filling in these concept representations. For example, one of the candidate lexicon entry for {\it shiny} is syntactic type: $\textit{objset}$, semantic program: $\textit{filter}(\textit{scene}(), \textbf{SHINY}_{\text{ObjConcept}})$.  During training, all these vector embeddings as well as the weights associated with each lexicon entry, will be optimized jointly.

Next, we discuss the implementation for two conjunctive lexicon entries. The grammar rule for $\text{CONJ}_{\text{AND}}$ is:
\[T~~\text{CONJ}_{\text{AND}}~~T \rightarrow T,\]
where $T$ is an arbitrary syntactic type (thus called generalized conjunction). There are two typical use cases: {what is the shape of the \underline{red and shiny} object}, and {what is the shape of the object that is \underline{left of the cube and right of the sphere}.} In the first case, both arguments have syntactic type $\textit{objset}/\textit{objset}$. In the second case, both arguments have syntactic type $\textit{objset}\backslash\textit{objset}$. Note that CLEVR contains only the second case.

The grammar rule for $\text{CONJ}_{\text{OR}}$ is:
\[\textit{objset}~~\text{CONJ}_{\text{OR}}~~\textit{objset} \rightarrow \textit{objset}\backslash\textit{objset}.\]
It covers the case: {\it how many objects are \underline{blue cubes or red spheres}.} Our implementation is slightly different with human-defined lexicon entries for the word {\it or}, in particular, because the DSL we use is a small set of set-theoretic operations, which does not fully match the expressiveness of truth-conditional semantics. Thus, the current DSL does not support the representation of all words in the dataset (in particular, {\it or} and {\it are}). Thus, we have implemented this ad-hoc fix to handle disjunction.

Finally, we want to emphasize again that, since our DSL does not support representing all semantic programs of words, we allow certain words to be associated with an ``{\it empty}'' lexicon entry. This entry can be combined with any words or constituents next to it and does not participate in the composition of syntactic types and semantic programs. In \tbl{tab:clevr-lexicon-example} we show the lexicon entry associated with each word in the sentence ``{\it are there any shiny cubes?}'', learned by our model, \model.

\begin{table}[t]
    \centering
    \setlength{\tabcolsep}{12pt}
    \begin{tabular}{lll} \toprule
        Word Type & Syntactic Type & Semantic Program \\ \midrule
        are & <EMPTY> & <EMPTY>\\
        there & <EMPTY> & <EMPTY>\\
        any & $ \textit{bool}/\textit{objset}$ & $\lambda x.\textit{exist}(x)$\\
        shiny & $\textit{objset}/\textit{objset}$ & $\lambda x.\textit{filter}(x, \textbf{SHINY}_{\text{ObjConcept}})$\\
        cubes & $\textit{objset}$ & $\textit{filter}(\textit{scene}(), \textbf{CUBE}_{\text{ObjConcept}})$\\
    \bottomrule
    \end{tabular}
    \vspace{5pt}
    \caption{The learned lexicon entries associated with each word for a simple sentence: {\it are there any shiny cubes?}. The derived semantic program for the full sentence is $\textit{exist}(\textit{filter}(\textit{filter}(\textit{scene}(), \textbf{CUBE}_{\text{ObjConcept}}), \textbf{SHINY}_{\text{ObjConcept}}))$}
    \label{tab:clevr-lexicon-example}
\end{table}

\subsection{Language-Driven Navigation DSL}

Our DSL for the language-driven navigation domain is a simple string manipulation language that supports creating new strings, concatenating two strings, and repeating a string multiple times. Our DSL contains only two primitive types: action sequence, abbreviated as ActSeq, and integer.

Formally, we summarize the list of operations in our language-driven navigation domain in \tbl{tab:scan-dsl}.

\begin{table}[tp]
    \centering
    \setlength{\tabcolsep}{3pt}
    \begin{tabular}{p{0.45\columnwidth}p{0.5\columnwidth}} \toprule
        Signature & Note\\ \midrule
        {\it empty}() $\longrightarrow$ ActSeq & Create an empty string (of length 0).\\ \midrule
        {\it newprim}() $\longrightarrow$ ActSeq & Create a string containing only one primitive action. In SCAN, there are in total 6 primitives.\\ \midrule
        {\it newint}() $\longrightarrow$ Integer & Create a single integer. In SCAN, we only support integers \{2, 3, 4\}.\\ \midrule
        {\it concat}($\mathbf{a}$: ActSeq, $c$: ActSeq) $\longrightarrow$ ActSeq & Concatenate two input strings. \\ \midrule
        {\it repeat}($\mathbf{a}$: ActSeq, $\mathbf{b}$: Integer) $\longrightarrow$ ActSeq & Repeat the input string for multiple times. \\\bottomrule
    \end{tabular}
    \vspace{5pt}
    \caption{All operations in the domain-specific language for language-driven navigation.}
    \label{tab:scan-dsl}
\end{table}

\paragraph{Probabilistic string representation.}
We represent each string in a ``probabilistic'' manner. In particular, each string $s$ is represented as a tuple $\langle L^s, C^s\rangle$. $L^s$ is a categorical distribution of the length. $C^s$ is a three-dimensional tensor, indexed by $\ell, k, c$, where $C^s_{\ell, k, c} = p(s[k] = c | \textit{length}(s) = \ell)$. Thus, $C$ has the shape $[L+1, L, |V|]$, where $L$ is the max length of a string and $V$ is the action vocabulary. For simplicity, we constrain that $C^s_{\ell, k, c} \equiv 0$ for all $k > \ell$.

It is straightforward to represent empty strings: $L_0 = 1$, or strings with a single action primitive $a$: $L_1 = 1$ and $C_{1, 0, a} = 1$. Now we explain our implementation of the $\textit{concat}$ and the $\textit{repeat}$ operation.

For $z = \textit{concat}(x, y)$:
\begin{eqnarray*}
L^z_\ell &=& \sum_{0 \le i \le \ell} \left( L^x_i \cdot L^y_{(\ell - i)} \right);\\
C^z_{\ell, k, c} &=& \frac{1}{L^z_\ell} \sum_{0 \le i \le \ell} \left( L^x_i \cdot L^y_{(\ell - i)} \cdot (C^x_{i, k, c} + C^y_{\ell - i, k - i, c}) \right).
\end{eqnarray*}
The high-level idea is to enumerate the possible length of both strings.

Similarly, for $z = \textit{repeat}(x, m)$,
\begin{eqnarray*}
L^z_\ell &=& \begin{cases}
    L^x_{\ell / m} & \text{if $\ell \text{~mod~} m = 0$}\\
    0 & \text{otherwise}
\end{cases}\\
C^z_{\ell, k, c} &=& \begin{cases}
    L^x_{\ell / m, k \text{~mod~} (\ell / m), c} & \text{if $\ell \mod m = 0$ and $k < \ell$}\\
    0 & \text{otherwise}
\end{cases}.
\end{eqnarray*}

\paragraph{Expected execution.} In the language-driven navigation domain, we only perform expected execution for semantic programs of type ActSeq, whose execution results can be represented using the probabilistic string representation. Denote $\bar{s}_i$ as the execution results for $K$ programs, and $\tau(s_i)$ the corresponding weights. We define $p(s_i) = \textit{softmax}\left( \left\{\tau(s_i)\right\} \right)_i = \frac{\exp \tau(s_i)}{ \sum_j \exp \tau(s)j) }$. We compute the expected string $\bar{s}$ and its weight $\tau(s)$ as:
\begin{equation*}
L^s_\ell = \sum_i p(s_i) L^{s_i}_\ell; \quad C^s_{\ell, k, c}  = \frac{ \sum_i \left( p(s_i) L^{s_i}_\ell \cdot C^{s_i}_{\ell, k, c} \right) }{L^s_{\ell}}. 
\end{equation*}

\paragraph{Candidate lexicons.}
We use a simple enumerate algorithm to generate candidate lexicon entries for our language-driven navigation DSL. Specifically, we first enumerate candidate semantic programs for each lexicon entry that satisfy the following constraints:
\begin{enumerate}
    \item There are at most three operations.
    \item There are at most two arguments.
    \item There is at most one argument whose type is a functor.
    \item There is no argument of type \textit{Integer}.
\end{enumerate}

\tbl{tab:scan-program-sample} lists a couple of programs generated by the algorithm and their corresponding types.

\begin{table}[t]
    \centering\small
    \setlength{\tabcolsep}{3pt}
    \begin{tabular}{lp{0.6\linewidth}} \toprule
        Type & Program (Note)\\ \midrule
        \mycell{ActSeq} & \mycell{$\textit{walk}()$ \\ The simplest program that constructs a string with a single action\\ primitive: {\bf WALK}.}\\ \midrule
        \mycell{(ActSeq) $\longrightarrow$ ActSeq} & \mycell{$\lambda x.\textit{concat}(\textit{look}(), x)$ \\ Prepend a {\bf LOOK} action to an input string.}\\ \midrule
        \mycell{(ActSeq, ActSeq) $\longrightarrow$ ActSeq} & \mycell{$\lambda x.\lambda y.\textit{concat}(x, y)$ \\ Concatenate two strings.}\\ \midrule
        \mycell{(ActSeq, ActSeq) $\longrightarrow$ ActSeq} & \mycell{$\lambda x.\lambda y.\textit{concat}(\textit{repeat}(x, 2), y)$ \\ Repeat the first string twice and concatenate with the second string.}\\ \midrule
        \mycell{((ActSeq) -> ActSeq, ActSeq) $\longrightarrow$ ActSeq} & \mycell{$\lambda x.\lambda y.\textit{concat}(y, x(\textit{walk}()))$ \\ The first argument ($x$) is a function which maps a ActSeq to\\ another ActSeq. The second argument $y$ is an ActSeq.\\ The function invokes $x$ with a simple string {\bf WALK}, and\\ concatenate the result with $y$.}\\
        \bottomrule
    \end{tabular}
    \vspace{5pt}
    \caption{Sample semantic programs generated by the enumeration process based on our language-driven navigation DSL.}
    \label{tab:scan-program-sample}
\end{table}

Based on the candidate semantic types, we first instantiate candidate lexicon entries by enumerate possible ordering (linearization) of the arguments. For example, the simple program $\lambda x.\textit{concat}(\textit{look}(), x)$ has two possible linearizations: {\it ActSeq}/{\it ActSeq} and {\it ActSeq}\textbackslash{\it ActSeq}. As discussed in the main paper, in order to handle parsing ambiguities, we further introduce two finer-grained syntactic types for the {\it ActSeq} type: {\it S} and {\it V}. In practice, we only allow the following set of syntactic types: {\it V}, {\it V/V}, {\it V\textbackslash V}, {\it V\textbackslash V/V}, {\it V\textbackslash V/(V\textbackslash V)}, and {\it S\textbackslash V/V}. In total, we have 178 candidate lexicon entries for each word.
\clearpage
\section{Delve Into Expected Execution}
\label{app:ckyee}

In this section, we will run a concrete example in a small arithmetic domain to demonstrate the idea of expected execution. Following that, we will prove an important invariance property that has guided our realization of different functional modules in both domains.

\subsection{\ckyee In An Arithmetic Domain}
\label{sec:2-1-example}
In this section, we will consider parsing a very simple sentence in an arithmetic domain. We will be using numbers and two arithmetic operations: $+$ and $\times$. Each number in the domain will be represented as a real value.

Suppose we have the following lexicon entries associated with three words, illustrated in \tbl{tab:arith-lexicon}. There are 4 candidate derivations of the sentence ``{\it ONE PLUS\_ONE MUL\_THREE}'', as illustrated in \tbl{tab:arith-derivation}. For simplicity, we will show the $\exp$ of weights. Thus, the probability of a derivation is proportional to the product of all lexicon entry weights.

\begin{table}[t]
    \centering\small
    \setlength{\tabcolsep}{12pt}
    \begin{tabular}{llll} \toprule
        Word Type & Syntactic Type & Semantic Program & $\exp$(Weight) \\ \midrule
        ONE & N & 1 & 1.0\\ \midrule
        PLUS\_ONE & N\textbackslash N & $\lambda x. x + 1$ & 0.5\\
        PLUS\_ONE & N\textbackslash N & $\lambda x. x \times 3$ & 0.5\\ \midrule
        MUL\_THREE & N\textbackslash N & $\lambda x. x + 1$ & 0.5\\
        MUL\_THREE & N\textbackslash N & $\lambda x. x \times 3$ & 0.5\\
    \bottomrule
    \end{tabular}
    \vspace{5pt}
    \caption{A set of candidate lexicon entries and their weights in a simple arithmetic domain.}
    \label{tab:arith-lexicon}
\end{table}

\begin{table}[t]
    \centering\small
    \setlength{\tabcolsep}{12pt}
    \begin{tabular}{lllll} \toprule
        \mycell{Index} & \mycell{Syn. Type} & \mycell{Semantic Program\\ ($\textit{derivation}_k$)} & \mycell{Execution Result\\ ($\textit{exec}_k$)} & \mycell{$\exp$(Weight)} \\ \midrule
        1 & N & $(1 + 1) + 1$ & $3$ & $0.25 = 1 \times 0.5 \times 0.5$\\
        2 & N & $(1 \times 3) + 1$ & $4$ & $0.25 = 1 \times 0.5 \times 0.5$\\
        3 & N & $(1 + 1) \times 3$ & $6$ & $0.25 = 1 \times 0.5 \times 0.5$\\
        4 & N & $(1 \times 3) \times 3$ & $9$ & $0.25 = 1 \times 0.5 \times 0.5$\\
    \bottomrule
    \end{tabular}
    \vspace{5pt}
    \caption{Four candidate derivations of the simple sentence ``{\it ONE PLUS\_ONE MUL\_THREE}'' in a simple arithmetic domain.}
    \label{tab:arith-derivation}
\end{table}

Suppose that we will be using the groundtruth execution result of this program as the supervision, applied by an L2 loss. Then we will be interested in the expected execution result of all possible derivations. In this case, it is
\[ \mathbb{E}[\textit{exec}] = 0.25 \times 3 + 0.25 \times 4 + 0.25 \times 6 + 0.25 \times 9 = 5.5. \]

Next, we will try to accelerate the computation of $\mathbb{E}[\textit{exec}]$ by doing local marginalization. Consider the constituent ``{\it ONE PLUS\_ONE}''. In \ckyee, this will be the first constituent that the algorithm constructs. It has two possible derivations, whose corresponding semantic programs are $(1 + 1)$ and $(1 \times 3)$. Both derivations have the same syntactic type N, and thus, they will be combined with N\textbackslash N on its right, in the next step. In this case, in \ckyee, we will merge these two derivations into one (again, since we only care about the expected execution result, not the set of all possible derivations!). The combined derivation has value $0.5 \times (1 + 1) + 0.5 \times (1 \times 3) = 2.5$, and total weight $0.5 + 0.5 = 1$.

Then, when we are trying to compose the derivation for the whole sentence, \ie, combine the constituent ``{\it ONE PLUS\_ONE}'' and ``{\it MUL\_THREE}'', we no longer need to compute all $2\times 2 = 4$ possible derivations, but only $1 \times 2 = 2$ derivations. They are: $2.5 + 1 = 3.5$, with probability $0.5$, and $2.5 \times 3 = 7.5$, with probability $0.5$. In this case, we see that taking local marginalization reduces the computation complexity of parsing and retains the expected execution result!

\subsection{Formal Properties of \ckyee}
\label{sec:2-2-proof}
Motivated by the intuitive example shown above, now let us formally specify the properties of \ckyee.

\paragraph{Expectation invariance.} Consider the composition of two consecutive constituents $a$ and $b$. We use $a_1, \cdots a_N$ and $b_1, \cdots b_N$ to denote possible derivations of both constituents. We assume all $a_i$'s are of the same syntactic type without loss of generality, so do all $b_i$'s, since we will handle different syntactic types separately.

Denote $f$ as the semantic composition function for $a$ and $b$. Without local marginalization, we will have in total $N \times M$ derivations for the result constituent: $c_{i,j} = f(a_i, b_j)$. We further use $\bar{a}_i$, $\bar{b}_j$, and $\bar{c}_{i,j}$ to denote the execution results of these derivations. Without derivations, the expected execution results is:

\begin{eqnarray*}
\mathbb{E}[\bar{c}] &=& \frac{1}{\sum_{i,j} \exp\tau(c_{i,j})} \sum_{i,j} \left( \exp\tau(c_{i,j}) \cdot \bar{c}_{i,j} \right)\\
&=& \frac{1}{\sum_{i,j} \exp\tau(a_i) \cdot \exp \tau(b_j)} \sum_{i,j} \left( \exp\tau(a_i) \cdot \exp \tau(b_j) \cdot f(\bar{a}_i, \bar{b}_j) \right) .\\
\end{eqnarray*}

Again, without loss of generality, we will assume $\sum_i \exp\tau(a_i) = 1$ and $\sum_j \exp\tau(b_j) = 1$, because any constant scaling of these weights will not change the expectation $\mathbb{E}(\bar{c})$. Thus, we simplify this definition as,
\[ \mathbb{E}[\bar{c}] = \sum_{i,j} \left( \exp\tau(a_i) \cdot \exp \tau(b_j) \cdot f(\bar{a}_i, \bar{b}_j) \right).\]
Let us assume function $f$ has the following property: $\mathbb{E}[f(a, b)] = f(\mathbb{E}(a), \mathbb{E}(b))$, which expands as,
\[ \sum_{i,j} \left( \exp\tau(a_i) \cdot \exp\tau(b_j) f(\bar{a}_i, \bar{b}_j)\right) = f\left(\sum_{i}\left( \exp\tau(a_i) \bar{a}_i \right), \sum_{j}\left( \exp\tau(b_j) \bar{b}_j \right) \right). \]
Thus, locally marginalizing the expected value for $\mathbb{E}[\bar{a}]$ and $\mathbb{E}[\bar{b}]$ will not change the expected execution result of $c$.

In this simple proof we have made a strong assumption on the composition function $f$: $\mathbb{E}[f(a, b)] = f(\mathbb{E}(a), \mathbb{E}(b))$. In practice, this is true when $f$ are addition or multiplication functions between scalars, vectors, matrices, and in general, tensors. It will not apply to element-wise min/max operations and other non-linear transformations. Fortunately, this already covers most of the operations we use in the visual reasoning and language-driven navigation DSLs. In practice, even if some operations do not have this property, we can still use this mechanism to approximate the expected execution result.

Although we have only proved this property for binary functions, the idea itself easily generalizes to unary functions, such as the negation operation, and higher-arity functions. Furthermore, by induction over derivation trees, we can easily prove that, as long as all composition functions satisfy the expectation invariance property, applying \ckyee yields the same result as doing the marginalization at a sentence level.

\paragraph{Complexity.} In general, it is hard to quantify the reduction in computational complexity by doing local marginalization. However, we can still estimate the number of possible derivations constructed in the entire \ckyee procedure. For simplicity, consider the case where there is only one primitive syntactic type: {\it X}. Moreover, there are $N_0$ candidate lexicon entries of type {\it X}; $N_1$ entries of type {\it X/X}, and $N_2$ entries of type {\it X\textbackslash X/X}. For each span, considered in the \ckyee algorithm, all possible derivations associated with this span can be grouped into 4 categories:
\begin{enumerate}
    \item Derivations of type {\it X}. In this case, only 1 derivation will be retained (merged by the expected execution result).
    \item Derivations of type {\it X/X}. In this case, they must be a primitive lexicon entry. Thus, there are at most $N_1$ of them.
    \item Derivations of type {\it X\textbackslash X/X}. Similarly, at most $N_2$ of them.
    \item Derivations of type {\it X\textbackslash X}. This intermediate syntactic type is a result of a partial composition between {\it X\textbackslash X/X} and {\it X} (on its right). Thus, there are at most $N_2$ of them.
\end{enumerate}
Overall, there are at most $1 + N_1 + 2\times N_2$ derivations stored for this span. Since the total span is $O(L^2)$, where $L$ is the total length of the input sentence, the overall complexity of \ckyee is a polynomial of $L$, $N_0$, $N_1$, and $N_2$, which is significantly lower than an exponential number of derivations.

\subsection{Connection with Other Parsing Models}

\paragraph{Connection with synchronous grammars.} Our approach can be viewed as defining a synchronous grammar over joint (parse tree, meaning program) pairings. We didn’t use this framing in the main paper because typical applications of synchronous grammar involve parallel datasets (\eg, sentence pairs in two languages for machine translation, or sentence–image pairs for generating image descriptions) in which the information in both modalities is directly parsed. In our setting, in contrast, the meaning-program component of the synchronous grammar is acquired through more distant supervision. We will make all this clear in the final version, including stating how the chart parsing process can be seen as synchronously constructing parsing trees and meaning programs. The expected execution can be viewed as a ``compression'' step over all meaning programs that can be potentially parsed from each span.

\paragraph{Connection with sum-product CKY.} There are also connections between \ckyee and sum-product CKY, such as the shared Markovian assumption, but would like to add that the main difference between \ckyee and sum-product CKY is that \ckyee computes only the “expectation” of the execution results of the underlying program. Instead, sum-product CKY computes a full distribution of the parsing results (e.g., in syntax parsing, it can compute the categorical distribution of the root symbol). Sum-Product CKY can not be applied to our setting, because we are dealing with programs and the space of possible programs is infinite. Modeling a distribution of all possible programs might be intractable; instead, computing the expectation is much easier, and this enables us to do local marginalization.
\clearpage

\section{Experimental Setup And Analysis}
\label{sec:3-datasets}
In this section, we will present in detail the experimental setups for both datasets: CLEVR and SCAN. Both datasets are released under a BSD license. Specifically, we will include details about the setups for different compositional generalization tests. Although some of them (the ones in the SCAN dataset) have already been illustrated in their original paper, we echo them here for completeness. Next, we will also analyze the performance of each model on both datasets, focusing on the inductive biases they have in their model and how these inductive biases contribute to their compositional generalization in different splits. 

\subsection{Visual Reasoning: CLEVR}
We start with the dataset generation process for the CLEVR dataset. Next, we formally present the dataset generation protocol for all splits. Furthermore, we analyze the performance of various models.

\paragraph{Baselines.} As a quick recap of our baselines, MAC~\cite{Hudson2018Compositional} uses an end-to-end vision-language attention mechanism to process the question and image jointly; TbD-Nets~\cite{Mascharka2018Transparency} and NS-VQA~\cite{Yi2018Neuro} uses a neural sequence-to-sequence model (with attention, see~\cite{Bahdanau2015Neural} for semantic parsing. The parser is trained with sentence-program pairs; NS-CL~\cite{Mao2019NeuroSymbolic} uses a customized sequence-to-tree model, and jointly learns the visual recognition models and the semantic parser. One crucial implementation detail with the semantic parser module in NS-CL is that, it uses additional token embeddings to annotate the concepts appearing in the question. When generating a concept token in the semantic program, it uses an attention mechanism to select the concept from the input question.

\paragraph{Dataset generation.} Since we only consider the cases where each word is associated with a unique lexicon entry, we have manually excluded sentences that will break this assumption. Among all of the 425 templates in the original CLEVR dataset~\cite{Johnson2017CLEVR}, we have retained 208 templates. Specifically, we have removed all templates that involve: 1) coreference resolution, 2) ``same''-related questions, and 3) number comparison-related questions. To keep the number of questions the same as the original dataset, we choose to re-generate the questions using the selected subset of templates, following the original data generation protocol. All our splits are generated based on this basic version, which we name as the standard training set and the standard test set.

\paragraph{Split: data efficiency.}

We test the data efficiency of models by only using 10\% of training data in the standard training set, and test the models on the standard test set.

In this split, the semantic parsers used by all program-based methods: TbD-Nets, NS-VQA, and NS-CL, have nearly perfect accuracy. Thus, the performance drops are primarily due to the limited data for training individual modules. Overall, TbD-Nets have the worst data efficiency. There is no performance drop for the NS-VQA model, because the visual recognition modules are pretrained with direct object-level supervision.

\paragraph{Split: compositional generalization (purple).}
The training set is generated by selecting all questions that either do not contain the word ``purple'' or have a length smaller than or equal to 7 (including punctuation). The test set is generated by selecting all sentences containing the word ``purple'' and has a length greater than 7.

In this split, the semantic parsers used by all program-based methods: TbD-Nets, NS-VQA, and NS-CL have nearly perfect accuracy. Thus, the performance drops are primarily due to 1) the limited data for training individual modules (in this case, {\it filter}({\it purple}) and 2) novel composition of learned modules. Overall, NS-VQA and NS-CL answer more questions correctly than TbD-Nets. 

\paragraph{Split: compositional generalization (right).}
The training set is generated by selecting all questions that either do not contain the word ``right'' or have a length smaller than or equal to 12. The test set is generated by selecting all sentences that contain the word ``right'' and have a length greater than 12.

In this split, the semantic parser of NS-CL still yields almost perfect accuracy. In contrast, the accuracy of the neural sequence-to-sequence parser used by TbD-Nets and NS-VQA is around 91\%. Thus, the performance drop of TbD-Nets is mainly due to the inferior performance of the corresponding neural module: {\it relate}({\it right}). Compared with the realization of a {\it filter} module (used in our {\it purple} generalization test), the {\it relate} module in TbD-Nets has a significantly deeper neural architecture (6 layers vs. 3 layer). Thus, we hypothesize that this module requires more data to train.

\paragraph{Split: compositional generalization (count).}
The training set is generated by selecting all questions that either do not contain operation ``count'' or have a length smaller than or equal to 9. The test set is generated by selecting all sentences that contain operation ``count'' and have a length greater than 9.

Among all compositional generalization tests, this is the most challenging one. The semantic parser, in this case, need to generalize from: ``{\it how many cubes are there}'' and ``{\it what's the shape of the object that is both left of the cube and right of the sphere?}'', to ``{\it how many cubes are both left of the cube and right of the sphere?}'' We have constructed the training data in a way such that all constituents have been seen in the training data. In this test, the program-level accuracy of the semantic parser used by TbD-Nets and NS-VQA is 70.8\%. NS-VQA outputs slightly higher QA accuracy.

We find that, for the parsing module in NS-CL, it fails to output the correct filter operation. Given the input question ``what is the number of spheres that are right of the cube'', it sometimes outputs {\it filter\_cube relate\_right filter\_cube count} (the correct program has the third operation {\it filter\_sphere} instead). This is because the system has never seen sentences composed of “counting” operations and such complex structures; during training, it has only seen short sentences such as ``what is the number of spheres?'' Only our G2L2 model, with its explicit constituent-level compositionality, is able to make these generalizations.

\paragraph{Split: depth generalization.}
We define the ``hop number'' of a question as the number of intermediate objects being referred to in order to locate the target object. For example, the ``hop number'' of the question ``{{\it how many red objects are right of the cube?}}'' is 1. We train different models on 0-hop and 1-hop questions and test them on 2-hop questions.

This generalization test evaluates the generalization to deeper syntactic structures. All methods except for our model \model fail on this test. By evaluating the accuracy of the program generated by different semantic parsers, we find that, the neural sequence-to-sequence model used by TbD-Nets and NS-VQA completely fails on this task, sometimes generating invalid programs (the program-level accuracy is 1.7\%). Thus, we see a significant performance drop for both methods. Meanwhile, NS-CL also generates wrong programs, but the programs are always valid due to its sequence-to-tree design. Furthermore, even if the program is not correct, for example, they miss certain operations, the execution result may still lead to a correct answer. For example, as long as the semantic parser gets the outer-most filter operation (\ie, the last hop) correct, it is still possible to generate the correct answer.

\subsection{Language-Driven Navigation: SCAN}
The SCAN dataset \citep{Lake2018Generalization} consists of several official splits for generalizability evaluation. Following existing work, we evaluate on the splits corresponding to the generalization test of ``jump'' and ``around right''. In addition, we test the generalizability across different output lengths and the data efficiency of models. The performance is measured by exact match--based output accuracy. 

\paragraph{Split: data efficiency.}
The official ``simple'' split randomly samples 80\% among all possible example pairs as the training set, and leaves the others as the test set. We use the training set of the official simple split (available at \href{https://raw.githubusercontent.com/brendenlake/SCAN/master/simple_split/tasks_train_simple.txt}{[this url]}) as the entire training set, and test the data efficiency by using only 10\% of them. We sample the 10\% data uniformly for each input length. Both settings are tested in the official simple test split, which is available at \href{https://raw.githubusercontent.com/brendenlake/SCAN/master/simple_split/tasks_test_simple.txt}{[this url]}.

\paragraph{Split: compositional generalization (jump).}
The training split consists of \textit{jump} in isolation, \ie, the input is \textit{jump} while the ground-truth output is \text{I\_JUMP}, along with all other examples that do not contain \textit{jump}. The model is expected to recognize that \textit{jump} has the same syntactic category as other verbs such as \textit{run}, and does well on complicated instructions including \textit{jump}, \eg, mapping \textit{jump twice} to \text{I\_JUMP I\_JUMP}.
The training data is available at \href{https://raw.githubusercontent.com/brendenlake/SCAN/master/add_prim_split/tasks_train_addprim_jump.txt}{[this url]} and the test data is available at \href{https://raw.githubusercontent.com/brendenlake/SCAN/master/add_prim_split/tasks_test_addprim_jump.txt}{[this url]}.

\paragraph{Split: compositional generalization (around right).}
Similar to the ``jump'' test, the training set for the ``around-right'' test consists of all possible examples that do not contain \textit{around right} in their inputs, while the test set consists of all examples that have \textit{around right}. It is worth noting that different from \textit{jump}, \textit{around right} in isolation is not a valid input as it lacks a primitive. The model is expected to perform compositional generalization, understanding \textit{around right} based on existing training phrases such as \textit{around left} and \textit{opposite right}. 
The training data is available at \href{https://raw.githubusercontent.com/brendenlake/SCAN/master/template_split/tasks_train_template_around_right.txt}{[this url]} and the test data is available at \href{https://raw.githubusercontent.com/brendenlake/SCAN/master/template_split/tasks_test_template_around_right.txt}{[this url]}.

\paragraph{Split: length generalization.} 
The model is expected to perform well on examples with long ground-truth output while training those with short ground-truth output. In this test, all training examples consist of less than or equal to 22 tokens in their outputs, while the output of a test example may consist of up to 48 tokens. 
The training data is available at \href{https://raw.githubusercontent.com/brendenlake/SCAN/master/length_split/tasks_train_length.txt}{[this url]} and the test data is available at \href{https://raw.githubusercontent.com/brendenlake/SCAN/master/length_split/tasks_test_length.txt}{[this url]}.

\paragraph{Baseline models.} All baseline models are built on top of a seq2seq model \citep{Sutskever2014Sequence}: the original seq2seq model \citep{Sutskever2014Sequence} trains an LSTM-based encoder-decoder model using the training set; the other methods augment the training set by either heuristics or learned models, and train an LSTM-based encoder-decoder model using the augmented data. It is worth noting that among all considered baseline methods, GECA \citep{andreas-2020-good} may generate examples in the test set, especially for compositional generalization tests since the heuristics it introduces is by nature compositional.  %

\end{document}